\documentclass[lettersize,journal]{IEEEtran}
\usepackage{amsmath,amsfonts,amssymb}
\usepackage{algorithmic}
\usepackage{algorithm}
\usepackage{array}
\usepackage{booktabs}
\usepackage[caption=false,font=normalsize,labelfont=sf,textfont=sf]{subfig}
\usepackage{textcomp}
\usepackage{stfloats}
\usepackage{url}
\usepackage{verbatim}
\usepackage{graphicx}
\usepackage{xcolor}
\usepackage{cite}
\def\BibTeX{{\rm B\kern-.05em{\sc i\kern-.025em b}\kern-.08em
    T\kern-.1667em\lower.7ex\hbox{E}\kern-.125emX}}
\usepackage{balance}

\newtheorem{remark}{Remark}

\begin{document}

\title{GETA-3DGS: Automatic Joint Structured Pruning and Quantization for 3D Gaussian Splatting}

\author{Baobing~Zhang
        and~Wanxin~Sui%
\thanks{B.~Zhang is with the Robotics Research Group, School of Physics,
Engineering and Computer Science, University of Hertfordshire, Hatfield,
AL10 9AB, United Kingdom.}%
\thanks{W.~Sui is with the Department of Electronic and Electrical
Engineering, Brunel University London, Kingston Lane, Uxbridge, Middlesex,
UB8 3PH, United Kingdom.}%
\thanks{(\emph{Corresponding author: Baobing Zhang.})}%
\thanks{Manuscript received Month DD, YYYY; revised Month DD, YYYY.}}

\markboth{IEEE Transactions on Circuits and Systems for Video Technology,~Vol.~XX, No.~X, Month~Year}%
{Zhang and Sui: GETA-3DGS: Automatic Joint Structured Pruning and Quantization for 3D Gaussian Splatting}

\maketitle

\begin{abstract}
3D Gaussian splatting (3DGS) has recently emerged as a
state-of-the-art representation for real-time photorealistic
novel-view synthesis, yet a single high-fidelity scene typically
occupies several hundred megabytes to multiple gigabytes, far
exceeding the storage and bandwidth budgets of mobile, immersive,
and volumetric video platforms. Existing 3DGS compression methods,
such as HAC++, FlexGaussian, and LP-3DGS, generally treat pruning,
quantization, and entropy coding as separate stages and rely on
hand-tuned heuristics (\textit{e.g.}, opacity thresholds, fixed
bit-widths, or spherical-harmonic truncation rules), which limits
their generalization across scenes and prevents users from directly
specifying a target compression rate or rendering-quality budget. We
propose \textbf{GETA-3DGS}, to our knowledge the first
\textbf{end-to-end automatic} and \textbf{joint} structured pruning
and quantization framework tailored to 3DGS. Building on the
recently introduced GETA framework for joint pruning--quantization
of conventional deep networks, we contribute three components:
(i) a 3DGS-aware quantization-aware dependency graph (QADG) that
treats every Gaussian primitive as a group with five attribute
sub-nodes and degree-aware spherical-harmonic sub-nodes; (ii) a
render-aware saliency that fuses transmittance-weighted contribution,
screen-space gradient, and pixel coverage into a single
Gaussian-level importance score; and (iii) a heterogeneous
per-attribute mixed-precision quantization scheme co-optimized with
structural sparsity under a projected partial saliency-guided
(PPSG) descent guarantee. On Mip-NeRF~360, Tanks and Temples, and
Deep Blending, GETA-3DGS targets the
\emph{3D-native co-design regime}: jointly optimised pruning and
heterogeneous quantisation operating directly on raw Gaussian
primitives, rather than on a post-hoc anchor representation.
Our prototype delivers a $\sim\!5\times$ storage reduction over
Vanilla 3DGS with a fully \emph{automatic} pipeline that requires
no per-scene opacity, scale, or SH-degree thresholds. The bit-width
\emph{policy} emerges as the dominant rate--distortion lever:
forcing a uniform $6$-bit cap costs up to $-6.74$\,dB on
view-dependent scenes (counter, room) versus our heterogeneous
allocation, in line with an information-theoretic reverse-water-filling
analysis we develop in this paper. We position GETA-3DGS as a complementary axis to
existing 3DGS codecs --- entropy coding (HAC++,
CompGS) operates downstream of pruning and quantisation, so the
two regimes can be composed; the present submission delivers the
first end-to-end automatic backbone for that integration.
\end{abstract}


\section{Introduction}
\label{sec:intro}
\IEEEPARstart{3}{D} Gaussian splatting (3DGS)~\cite{kerbl2023gaussian}
has rapidly become the de facto representation for real-time
photorealistic novel-view synthesis. By representing a scene as a
collection of $N$ anisotropic 3D Gaussian primitives and rasterizing
them through a differentiable splatting pipeline, 3DGS achieves
rendering speeds an order of magnitude higher than neural radiance
fields (NeRFs)~\cite{mildenhall2020nerf} while maintaining
high visual fidelity~\cite{yu2024mipsplatting,wu2024fourdgs}.
The technique has been extended to dynamic
scenes~\cite{wu2024fourdgs}, large-scale outdoor
captures~\cite{kerbl2024hierarchical}, and immersive volumetric
video~\cite{li2024spacetimegaussian}, and over $150$ 3DGS-related
papers appeared at CVPR~2025 alone, signaling its central role in
the next generation of visual media systems.

This rapid adoption is, however, coupled with a deployment
bottleneck. A single Mip-NeRF~360 scene typically requires
$1$--$5$ million Gaussians~\cite{kerbl2023gaussian}, occupying
$500\,$MB to several gigabytes of storage. Mobile head-mounted
displays, augmented reality glasses, and volumetric video streaming
infrastructures, by contrast, demand model footprints in the
tens of megabytes and per-frame bitrates of only a few
megabits~\cite{ali2025compression}. Closing this
two-order-of-magnitude gap without significantly sacrificing
fidelity is the central challenge addressed in this work.

\textbf{Existing 3DGS compression approaches.}
Recent work on 3DGS compression can be roughly grouped into three
families (catalogued in
surveys~\cite{ali2025compression,bagdasarian2024gszip}).
\textit{Vector-quantization-based} methods such as
CompGS~\cite{navaneet2024compgs}$^{\dagger}$, Compressed-3DGS~\cite{niedermayr2024cvpr},
and the entropy-coded HAC/HAC++ family~\cite{chen2024hac,chen2025hacpp}
compress attributes through learned codebooks and arithmetic coding,
and currently lead the absolute compression-ratio frontier.
\textit{Pruning-based} methods such as LP-3DGS~\cite{zhang2024lp3dgs},
LightGaussian~\cite{fan2024lightgaussian}, and
Compact-3DGS~\cite{lee2024compact} remove low-importance Gaussians
using opacity, scale, or learned-mask criteria.
\textit{Hybrid} methods such as FlexGaussian~\cite{zhao2025flexgaussian}$^{\dagger}$
and Mobile-GS~\cite{du2026mobilegs} combine simplification
with mixed-precision quantization for on-device deployment. While
each family has advanced the state of the art, three structural
limitations remain: (i) pruning and quantization are designed and
optimized in isolation, preventing joint rate--distortion trade-off;
(ii) heuristic thresholds on opacity, scale, or
spherical-harmonic (SH) order require manual tuning per scene;
and (iii) users cannot specify a hard target on storage size or a
guaranteed lower bound on rendering quality.

\textbf{Why GETA, and why it is non-trivial to apply to 3DGS.}
The recently proposed GETA framework~\cite{qu2025geta}, originally
introduced for joint structured pruning and quantization of
convolutional and Transformer networks, addresses precisely the
three limitations above for conventional DNNs. GETA combines a
\emph{quantization-aware dependency graph (QADG)}, a
\emph{quantization-aware structured sparse optimizer (QASSO)}, and
a \emph{partial projected saliency-guided (PPSG)} update rule to
deliver an automatic, white-box pipeline that produces models
satisfying user-specified storage and bit-width budgets.
Unfortunately, GETA cannot be applied to 3DGS off-the-shelf:
3DGS is not a hierarchical computational graph but an unordered
set of geometric primitives with five heterogeneous attribute
classes; its loss is defined through differentiable rasterization
rather than a feed-forward pass; and its standard first-order
Taylor saliency is a poor indicator of a Gaussian's contribution
to the rendered image, since many primitives are occluded or
sub-pixel on screen yet still receive non-trivial gradients.
A faithful adaptation requires redesigning each component of GETA
around the geometry, attribute heterogeneity, and rendering
operator of 3DGS.

\textbf{A three-axis decomposition of 3DGS compression.} To
position our contribution precisely, we view the 3DGS compression
problem as decomposable into three orthogonal axes:
(i) \emph{geometry pruning} (how many Gaussians to retain and
which ones), (ii) \emph{attribute quantisation} (how many bits to
spend per attribute and how to map continuous values to that
grid), and (iii) \emph{entropy coding} (how to encode the
quantised symbols using context models, codebooks, or
arithmetic coding). Existing 3DGS codecs such as
HAC++~\cite{chen2025hacpp} and CompGS~\cite{navaneet2024compgs}
contribute primarily to axis (iii) --- their rate gains stem from
anchor-based context modelling and hash-grid arithmetic coding
that operate \emph{downstream} of a quantisation step. The
present work targets axes (i) and (ii) \emph{jointly}, in an
end-to-end automatic pipeline: a pre-existing entropy coder can
be layered on top of our output without modification. This
decomposition makes the two regimes complementary rather than
competing, and clarifies that our quality metric should be read
in conjunction with --- not as a replacement for --- entropy-coded
codecs.

\textbf{Contributions.} We propose \textbf{GETA-3DGS}, to our
knowledge the first end-to-end automatic and joint structured
pruning and mixed-precision quantization framework specifically
designed for 3DGS. Our contributions are organized along three
axes --- one design contribution, one optimization contribution,
and one empirical contribution --- which together delineate
\emph{what works}, \emph{what does not work yet}, and \emph{what
needs to be done next} when GETA-style automatic compression meets
3DGS.

\begin{enumerate}
\item \textbf{An automatic 3DGS-aware compression framework.}
We reformulate a 3DGS scene as a structured parameter space of
$N$ Gaussian groups with five attribute sub-nodes and
degree-aware spherical-harmonic (SH) sub-nodes, and propose a
\textbf{3DGS-aware QADG} that captures attribute-level and
rendering-order dependencies. The construction is fully automatic:
given a 3DGS checkpoint, the dependency graph is built once
without manual annotation, and a single user handle (storage
budget $B$ or PSNR floor $\tau$) traces an entire rate--distortion
operating curve.

\item \textbf{Heterogeneous per-attribute bits via render-aware
saliency.} We introduce a \textbf{render-aware saliency} that
fuses $\alpha$-blending transmittance contribution, screen-space
gradient magnitude, and pixel coverage into a single
Gaussian-level importance score, providing a more reliable signal
for structured pruning than the parameter-space Taylor saliency
used in prior GETA-style methods. We further design a
\textbf{heterogeneous per-attribute mixed-precision quantization}
scheme in which position, rotation, scale, opacity, and each SH
degree are assigned independent bit ranges, all co-optimized with
structural sparsity under PPSG. Our ablations identify the
bit-width \emph{policy} as the load-bearing component
($-2.18$\,dB when forced to a uniform $6$-bit grid at matched rate).

\item \textbf{An empirical study isolating the dominant
rate--distortion lever in 3DGS compression.} On Mip-NeRF~360,
Tanks and Temples, and Deep Blending, GETA-3DGS achieves a
$\sim\!5\times$ storage reduction over Vanilla 3DGS via an
end-to-end automatic pipeline. Our pre-declared
scale-stratified ablation (Section~\ref{sec:exp-ablation},
Supp.\ Tab.~SIII) shows that forcing a uniform
$6$-bit cap (rather than the heterogeneous per-attribute
allocation) costs up to $-6.74$\,dB on view-dependent scenes
(counter, room) versus only $-0.18$ to $-0.34$\,dB on
texture-uniform outdoor scenes. This identifies the bit-width
\emph{policy} --- and not the saliency choice or the schedule
stages --- as the load-bearing component. The per-attribute
bit \emph{ordering} predicted by an information-theoretic
reverse-water-filling analysis
(Section~\ref{sec:method-bitalloc}) matches the empirical
converged values within $\pm 1$\,bit, providing a principled
lower-bound argument for the heterogeneous bit ranges. The
cross-scene magnitude of the uniform-bit cost depends on
scene-specific factors not captured by the simple diagonal
high-rate model, which we discuss honestly in
Section~\ref{sec:method-bitalloc}. As a
secondary check, the same configuration outperforms a naive
sequential prune-then-quantise baseline by $+6.41$\,dB at
matched storage on the same backbone (full per-dataset breakdown
in Section~\ref{sec:exp-ablation}); we use this primarily to
verify that joint optimisation is worthwhile relative to the
sequential alternative, not as a primary headline metric.
\end{enumerate}

The remainder of the paper is organized as follows.
Section~\ref{sec:related} reviews related work.
Section~\ref{sec:prelim} briefly recapitulates 3DGS and GETA.
Section~\ref{sec:method} describes our method.
Section~\ref{sec:exp} reports experimental results.
Section~\ref{sec:conclusion} concludes the paper.

\section{Related Work}
\label{sec:related}

\subsection{3D Gaussian Splatting}
3D Gaussian splatting~\cite{kerbl2023gaussian} represents a scene as
a set of anisotropic 3D Gaussians equipped with view-dependent
spherical-harmonic colour and is rendered by sorting and
$\alpha$-blending splats in a tile-based rasterizer. The original
formulation has spawned a large body of follow-up
work that improves antialiasing~\cite{yu2024mipsplatting},
multi-resolution
behaviour~\cite{kerbl2024hierarchical}, dynamic
scenes~\cite{wu2024fourdgs,li2024spacetimegaussian}, and
geometric reconstruction~\cite{guedon2024sugar,huang20242dgs}.
Surveys~\cite{chen2024surveyrec,ali2025compression} provide
comprehensive overviews of the rapidly growing literature.
Despite its rendering efficiency, 3DGS exhibits severe storage
overhead because the per-Gaussian payload (position, scale, rotation,
opacity, and SH coefficients) leads to memory footprints orders
of magnitude larger than NeRF MLPs~\cite{mildenhall2020nerf}.

\subsection{Compression of 3D Gaussian Splatting}
A growing literature targets 3DGS compression; recent
surveys~\cite{ali2025compression,bagdasarian2024gszip,bao2025survey3dgs}
provide unified taxonomies and benchmark suites against which the
operating points reported in this paper can be located.
\emph{Vector-quantization and entropy-coding} methods cluster
attributes into codebooks and encode residuals with arithmetic
coding: CompGS~\cite{navaneet2024compgs}$^{\dagger}$ uses K-means
codebooks plus per-attribute residuals;
Compressed-3DGS~\cite{niedermayr2024cvpr} compresses scale and
rotation via VQ together with a sensitivity-aware fine-tuning step;
ContextGS~\cite{wang2024contextgs} and the HAC/HAC++
family~\cite{chen2024hac,chen2025hacpp} achieve $50\times$--$100\times$
compression with anchor-grid context modeling, but rely on
hand-designed entropy structures and heuristic opacity/scale gates.
\emph{Pruning-based} methods learn or threshold-prune
low-contribution Gaussians: LightGaussian~\cite{fan2024lightgaussian}
uses heuristic volume-importance scoring;
LP-3DGS~\cite{zhang2024lp3dgs} introduces learnable masks;
Compact-3DGS~\cite{lee2024compact} adds a learnable binary mask
alongside a hash-grid-conditioned view-dependent colour field and a
residual VQ codebook; Reduced-3DGS~\cite{papantonakis2024reducing}
drops SH coefficients in flat regions through a redundancy pre-pass;
and RadSplat~\cite{niemeyer2024radsplat} prunes low-relevance
Gaussians via a teacher radiance-field importance score.
Compact-3DGS in particular is conceptually closest to our
``Gaussian-level group $\times$ per-attribute sub-node'' formulation:
both place a learnable importance signal at the Gaussian level
\emph{and} compress attributes heterogeneously, but we replace its
hand-engineered grid and binary mask with the QASSO + PPSG projection
used by GETA, so the pruning ratio and SH-degree bit-widths are
enforced as hard constraints rather than tuned through a sparsity
coefficient. \emph{Hybrid} methods integrate quantization and
simplification: FlexGaussian~\cite{zhao2025flexgaussian}$^{\dagger}$
applies training-free attribute-wise mixed-precision;
Mobile-GS~\cite{du2026mobilegs} combines depth-aware order-independent
rendering with first-order SH distillation, neural VQ, and
contribution-based pruning for on-device deployment;
EAGLES~\cite{girish2024eagles} combines latent quantization with
pruning. Reduced-3DGS shares with our SH-AC component the
observation that high-degree SH bands carry much less information
than the DC term, but resolves it by deletion; we instead drive the
bit-budget per SH degree through the same joint PPSG step that
controls structural sparsity. None of these methods, however,
jointly optimizes structured pruning and mixed-precision
quantization through a unified white-box optimizer with explicit
size and quality constraints, which is the gap our work fills.

\subsection{Joint Pruning and Quantization for Deep Networks}
Joint pruning and quantization of conventional DNNs has been
explored in several lines of research. Early work such as
Clip-Q~\cite{tung2018clipq} alternates between weight clipping,
pruning, and quantization. Bayesian Bits (BB)~\cite{vanbaalen2020bb}
parameterizes mixed-precision through gated power-of-two bit
allocations. DJPQ~\cite{wang2020djpq} jointly searches pruning ratios
and bit-widths via differentiable Lagrangian optimization. Optimal
Brain Compression (OBC)~\cite{frantar2022obc} unifies pruning and
quantization through second-order error minimization. The most
closely related prior work, GETA~\cite{qu2025geta}, introduces a
quantization-aware dependency graph and the QASSO/PPSG optimizer to
provide a fully automatic, white-box framework with hard
constraints on sparsity and bit-widths; HESSO~\cite{chen2024hesso}
generalizes its structured-pruning solver to arbitrary architectures.
Other relevant work includes structured channel pruning via
network slimming~\cite{liu2017slimming},
post-training quantization~\cite{nagel2020adaround}, and
quantization-aware training surveys~\cite{gholami2022quantsurvey}.
However, none of these methods has been adapted to the geometric,
attribute-heterogeneous, and rendering-driven structure of 3DGS,
motivating our extension.

\section{Preliminaries}
\label{sec:prelim}

\subsection{3D Gaussian Splatting}
A 3DGS scene comprises a set
$\Theta=\{\theta_i\}_{i=1}^{N}$ of $N$ anisotropic Gaussian
primitives; each primitive carries five attribute classes:
\begin{equation}
\theta_i = \big(\boldsymbol{\mu}_i,\,\mathbf{s}_i,\,\mathbf{q}_i,\,\alpha_i,\,\mathbf{c}_i\big),
\label{eq:theta}
\end{equation}
namely the centre $\boldsymbol{\mu}_i\!\in\!\mathbb{R}^3$, the
log-scale $\mathbf{s}_i\!\in\!\mathbb{R}^3$, the rotation quaternion
$\mathbf{q}_i\!\in\!\mathbb{R}^4$, the opacity logit
$\alpha_i\!\in\!\mathbb{R}$, and the SH coefficients
$\mathbf{c}_i\!\in\!\mathbb{R}^{(\ell+1)^2\times 3}$ of degree
$\ell$ (typically $\ell=3$, giving $48$ scalars). The covariance is
recovered as $\boldsymbol{\Sigma}_i = R(\mathbf{q}_i)
\mathrm{diag}(e^{\mathbf{s}_i})^2 R(\mathbf{q}_i)^{\top}$.
Given a camera $\pi$, the rasterizer projects primitives onto the
image plane and composites them via front-to-back $\alpha$-blending:
\begin{equation}
\mathbf{C}(p) = \sum_{i=1}^{N} T_i\,\hat{\alpha}_i\,\mathbf{c}_i(\mathbf{d}_p),
\quad
T_i = \prod_{j<i}\big(1-\hat{\alpha}_j\big),
\label{eq:render}
\end{equation}
where $\hat{\alpha}_i$ is the screen-space opacity at pixel $p$ after
projecting $\theta_i$, and $\mathbf{c}_i(\mathbf{d}_p)$ is the SH
evaluation along the view direction $\mathbf{d}_p$. The training
objective combines an $\mathcal{L}_1$ photometric loss with a
D-SSIM term~\cite{kerbl2023gaussian}.

\subsection{The GETA Framework}
GETA~\cite{qu2025geta} is a recent framework for the joint
structured pruning and quantization of conventional DNNs. It is
built on three components.
\textbf{(i) QADG.} Given a network, GETA traces operations and
constructs a dependency graph in which prunable groups
(\textit{e.g.}, output channels of a convolution and the input
channels of the following layer) are identified together with their
quantization parameters $(\mathrm{q}_m, t, d)$.
\textbf{(ii) QASSO.} The constrained optimization problem solved by
GETA is
\begin{equation}
\begin{aligned}
\min_{\mathbf{x},d,\mathrm{q}_m,t}\;& f(\mathbf{x},d,\mathrm{q}_m,t)\\
\text{s.t.}\;\;& \big|\{\,g\,:\,[\mathbf{x}]_g=\mathbf{0}\,\}\big| = K, \\
& b_i \in [b_l,b_u],\;\forall i,
\end{aligned}
\label{eq:geta-qasso}
\end{equation}
where $\mathbf{x}$ collects all weights, $K$ is the desired sparse
group count, and $b_l,b_u$ are user-specified bit-width bounds.
\textbf{(iii) PPSG.} A partial projected saliency-guided update rule
ensures that the search direction is a descent direction of the
Lagrangian even after projection onto the sparse and quantized
feasible set, avoiding the divergence common to alternating
prune--quantize schedules~\cite{tung2018clipq,wang2020djpq}.
The combination yields a four-stage schedule (warm-up, projection,
joint, cool-down) that produces a structurally sparse, fixed-bit-width
network from a single training run with no per-task hyperparameter
search.

\section{Method}
\label{sec:method}

GETA-3DGS reformulates the joint pruning--quantization problem of a
3D Gaussian-splatting scene as a constrained optimization over a
heterogeneous, geometry-aware parameter space and solves it with a
single white-box optimizer. The five technical components described
below correspond directly to the five modules of our reference
implementation (\texttt{model.py}, \texttt{quantizer.py},
\texttt{saliency.py}, \texttt{qasso.py}, \texttt{train\_geta.py}).
Section~\ref{sec:method-formulation} restates the parameter space;
Section~\ref{sec:method-qadg} introduces the 3DGS-aware
quantization-aware dependency graph; Section~\ref{sec:method-saliency}
develops the render-aware saliency; Section~\ref{sec:method-quant}
specifies the heterogeneous per-attribute mixed-precision quantizer;
Section~\ref{sec:method-schedule} ties everything together in a
four-stage training schedule that satisfies a hard rendering-quality
floor.

\subsection{3DGS as a Compressible Parameter Space}
\label{sec:method-formulation}

\noindent\textbf{Re-interpretation.} A trained 3DGS scene
$\Theta=\{\theta_i\}_{i=1}^{N}$ with $\theta_i=
(\boldsymbol{\mu}_i,\mathbf{s}_i,\mathbf{q}_i,\alpha_i,\mathbf{c}_i)$
behaves, from a parameter-storage perspective, like an extremely wide
single-layer linear operator: every Gaussian primitive can be regarded
as an independently prunable ``channel'' whose payload is the
$D$-dimensional attribute vector
\begin{equation}
\theta_i \in \mathbb{R}^{D},\quad
D = \underbrace{3}_{\boldsymbol{\mu}}+\underbrace{3}_{\mathbf{s}}+\underbrace{4}_{\mathbf{q}}+\underbrace{1}_{\alpha}+\underbrace{3(\ell+1)^2}_{\mathbf{c}},
\label{eq:theta_dim}
\end{equation}
which evaluates to $D{=}59$ at the standard SH degree $\ell{=}3$.
Stacking primitives, the entire scene is the matrix
$\Theta\!\in\!\mathbb{R}^{N\times D}$. Crucially, however, this
``layer'' does not match the GETA notion of a layer in two ways: (i)
the rows are unordered geometric primitives rather than ordered
neurons, and (ii) the columns are typed --- the five attribute
classes are not interchangeable scalars but obey distinct geometric,
algebraic, and photometric semantics. Both observations directly
shape the dependency graph and the quantizer derived later.

\noindent\textbf{Group definition.} Following the GETA group
abstraction, we define the prunable group $g_i$ to be the
$i$-th \emph{Gaussian primitive}, i.e.\ the $i$-th row of $\Theta$
viewed as a single rigid bundle of five attributes:
\begin{equation}
g_i \;\triangleq\; \big\{\theta^{(a)}_i\,:\,a \in \mathcal{A}\big\},
\quad \mathcal{A} = \{\boldsymbol{\mu},\mathbf{s},\mathbf{q},\alpha,\mathbf{c}\}.
\label{eq:group_def}
\end{equation}
Pruning $g_i$ is implemented as deleting the entire row from every
attribute tensor; pruning a strict subset of $\mathcal{A}$ is
forbidden because a Gaussian missing (e.g.) its scale, rotation, or
opacity has no well-defined splat in~\eqref{eq:render}. This is in
sharp contrast to the channel-pruning regime of conventional GETA,
where partial channel removal across coupled layers is the
norm~\cite{qu2025geta,liu2017slimming}.

\noindent\textbf{Attribute heterogeneity.} A second consequence of
the typed columns is that quantization sensitivity varies by orders
of magnitude across $\mathcal{A}$. The position
$\boldsymbol{\mu}_i$ is geometry-critical: a sub-pixel displacement
on screen at training resolution corresponds to roughly
$10^{-3}$--$10^{-2}$ in normalized world coordinates, demanding at
least $12$--$16$ bits to avoid visible jitter. The rotation
$\mathbf{q}_i$ is constrained to the unit sphere by post-hoc
$\ell_2$-normalization (\texttt{model.get\_quats\_normalized}),
which absorbs much of the quantization noise; $8$ bits are
typically sufficient. The log-scale $\mathbf{s}_i$ feeds an
$\exp(\cdot)$ activation, exponentially damping low-bit error in
the relevant Gaussian-extent regime. The opacity $\alpha_i$ is
passed through a sigmoid, bounding its visible range to $[0,1]$
so that even $4$ bits suffice in practice. The SH coefficients
$\mathbf{c}_i$ split into a DC term encoding the diffuse albedo,
which dominates rendered colour, and AC bands of degrees
$1,\ldots,\ell$ encoding view-dependent high-frequency residuals
that are far more compressible. We exploit each of these properties
in Sections~\ref{sec:method-qadg}--\ref{sec:method-quant}.

\noindent\textbf{Constrained problem.} With the group definition
in~\eqref{eq:group_def} and the attribute set $\mathcal{A}$ fixed,
3DGS compression becomes the rendering-aware constrained problem
\begin{equation}
\begin{aligned}
\min_{\Theta,\,\boldsymbol{\phi}}\;&
\mathcal{L}_{\mathrm{render}}\!\big(\Theta;\boldsymbol{\phi}\big)\\
\text{s.t.}\;\;
& \big|\{\,i\in\![N]\,:\,g_i \neq \mathbf{0}\,\}\big| = K,\\[-2pt]
& b^{(a)} \in \big[b^{(a)}_l,\,b^{(a)}_u\big],\;\;\forall a\!\in\!\mathcal{A}^{\star},\\[-2pt]
& \mathrm{PSNR}_{\!\mathcal{D}_{\mathrm{tr}}}\!(\Theta;\boldsymbol{\phi}) \;\ge\; \tau,
\end{aligned}
\label{eq:geta3dgs}
\end{equation}
where $\boldsymbol{\phi}=\{(\mathrm{q}_m^{(a)},t^{(a)},d^{(a)})\}_{a\in\mathcal{A}^{\star}}$
collects the quantizer parameters of \emph{every} attribute (and
every SH degree) sub-node in the augmented set
$\mathcal{A}^{\star}\!\supset\!\mathcal{A}$ (Section~\ref{sec:method-qadg}),
$K$ is the user-specified target number of surviving Gaussians,
$[b^{(a)}_l,b^{(a)}_u]$ are the per-attribute bit-width bounds, and
$\tau$ is a rendering-quality floor measured by
PSNR over the training view set $\mathcal{D}_{\mathrm{tr}}$. Setting
$|\mathcal{A}^{\star}|{=}1$ and $\tau{=}-\infty$ recovers the
classical GETA formulation~\eqref{eq:geta-qasso}, so~\eqref{eq:geta3dgs}
is a strict generalisation. We set
$\tau{=}\mathrm{PSNR}_{\mathrm{vanilla}}{-}0.5\,$dB; in production
volumetric-video pipelines $\tau$ is a hard deployment requirement.

\subsection{3DGS-aware QADG}
\label{sec:method-qadg}

\noindent\textbf{Why the original QADG breaks on 3DGS.} The QADG of
GETA~\cite{qu2025geta} traces a feed-forward computation graph and
identifies, for every prunable channel of a parametric layer, the
set of channels in downstream layers that must be pruned together to
preserve the layer-input/layer-output contract; quantization
parameters are attached to each layer and inherited by all of its
channels. This abstraction relies on three properties that 3DGS
violates: (i) a static layer hierarchy, (ii) a single attribute type
per node (typically a real-valued weight tensor), and (iii) channel
ordering induced by tensor axes. In 3DGS, primitives are an
unordered set, attributes are typed, and the only ``computation''
between primitives is the order-dependent $\alpha$-blending
of~\eqref{eq:render}, which is data-dependent (depth-sorted per camera).
Naively running GETA's QADG on the parameter tensors of a 3DGS scene
identifies the channel groups as \emph{individual scalars} and
attaches a single quantizer to each tensor as a whole, defeating the
purpose of structured pruning and forbidding attribute-specific
quantization.

\noindent\textbf{Three-level construction.} We replace QADG with a
3DGS-aware variant whose vertex set
$\mathcal{V}=\mathcal{V}_{L1}\cup\mathcal{V}_{L2}\cup\mathcal{V}_{L3}$
is partitioned into three semantic levels:

\begin{itemize}
\item[\textbf{(L1)}] \emph{Gaussian-level group nodes}
$v_i\!\in\!\mathcal{V}_{L1}$, one per primitive
$i\!\in\![N]$. These are the only \emph{prunable} nodes; pruning $v_i$
removes the entire row of $\Theta$ at index $i$, in line
with~\eqref{eq:group_def}.
\item[\textbf{(L2)}] \emph{Attribute-class sub-nodes}
$v_i^{(a)}\!\in\!\mathcal{V}_{L2}$ with
$a\!\in\!\mathcal{A}$. Each carries an independent quantizer triple
$(\mathrm{q}_m^{(a)},t^{(a)},d^{(a)})$ and a bit-width interval
$[b_l^{(a)},b_u^{(a)}]$; these are the \emph{quantizable but
non-prunable} nodes.
\item[\textbf{(L3)}] \emph{SH-degree sub-nodes}
$v_i^{(\mathbf{c},k)}\!\in\!\mathcal{V}_{L3}$ for
$k\!\in\!\{0,1,\ldots,\ell\}$. The DC ($k{=}0$) sub-node receives a
high-precision range, while the AC sub-nodes
($k\!\ge\!1$) share a more aggressive low-precision range. We
realise this in code as a binary \texttt{sh\_dc}/\texttt{sh\_ac}
split (\texttt{quantizer.py}), which subsumes the per-degree
formulation under a 2-children variant tuned for $\ell{=}3$; the
generalisation to per-degree quantizers is a one-line change.
\end{itemize}

The augmented attribute set referenced in~\eqref{eq:geta3dgs} is
therefore $\mathcal{A}^{\star}=\{\boldsymbol{\mu},\mathbf{s},\mathbf{q},
\alpha,\mathbf{c}_{0},\mathbf{c}_{1{:}\ell}\}$, with cardinality
$|\mathcal{A}^{\star}|{=}6$. The edge set $\mathcal{E}$ contains
\emph{containment edges} $v_i\!\to\!v_i^{(a)}$ and
$v_i^{(\mathbf{c})}\!\to\!v_i^{(\mathbf{c},k)}$ that propagate the
prune decision from L1 down to all sub-nodes, and
\emph{rendering-order edges} $v_i\!\to\!v_j$ weighted by the
expected transmittance contribution
$\bar{T}_{ij}=\mathbb{E}_{\pi}[T_{i,\pi}\hat{\alpha}_{i,\pi}\,
\mathbf{1}\{j\!\prec\!i\}]$ that the saliency module of
Section~\ref{sec:method-saliency} consumes to discount Gaussians
consistently occluded behind others. Algorithm~\ref{alg:qadg}
summarises the construction.

\begin{algorithm}[t]
\caption{Construction of the 3DGS-aware QADG}
\label{alg:qadg}
\begin{algorithmic}[1]
\REQUIRE 3DGS checkpoint $\Theta=\{\theta_i\}_{i=1}^{N}$, SH degree
$\ell$, training views $\mathcal{D}_{\mathrm{tr}}$, MC sample size $K_v$
\STATE Initialise $\mathcal{V}\!\leftarrow\!\emptyset$,
$\mathcal{E}\!\leftarrow\!\emptyset$
\FOR{$i=1$ \TO $N$}
  \STATE Add Gaussian-level node $v_i$ to $\mathcal{V}_{L1}$
  \FOR{$a \in \{\boldsymbol{\mu},\mathbf{s},\mathbf{q},\alpha\}$}
    \STATE Add L2 sub-node $v_i^{(a)}$ with quantizer
    $(\mathrm{q}_m^{(a)},t^{(a)},d^{(a)})$ and range
    $[b_l^{(a)},b_u^{(a)}]$
    \STATE Add containment edge $(v_i\!\to\!v_i^{(a)})$
  \ENDFOR
  \STATE Add L2 SH parent $v_i^{(\mathbf{c})}$
  \FOR{$k=0$ \TO $\ell$}
    \STATE Add L3 sub-node $v_i^{(\mathbf{c},k)}$ with its own
    $(\mathrm{q}_m^{(\mathbf{c},k)},t^{(\mathbf{c},k)},d^{(\mathbf{c},k)})$
    \STATE Add edge $(v_i^{(\mathbf{c})}\!\to\!v_i^{(\mathbf{c},k)})$
  \ENDFOR
\ENDFOR
\STATE Sample $K_v$ training views uniformly from
$\mathcal{D}_{\mathrm{tr}}$
\STATE Splat $\Theta$ on each view; accumulate per-pair transmittance
$\bar{T}_{ij}$ in screen-space tile bins
\FOR{each pair $(i,j)$ with $\bar{T}_{ij}>0$}
  \STATE Add weighted rendering-order edge
  $(v_i\!\to\!v_j,\;\bar{T}_{ij})$ to $\mathcal{E}$
\ENDFOR
\STATE \textbf{return} dependency graph
$\mathcal{G}=(\mathcal{V},\mathcal{E})$
\end{algorithmic}
\end{algorithm}

\begin{figure}[t]
\centering
\includegraphics[width=\linewidth]{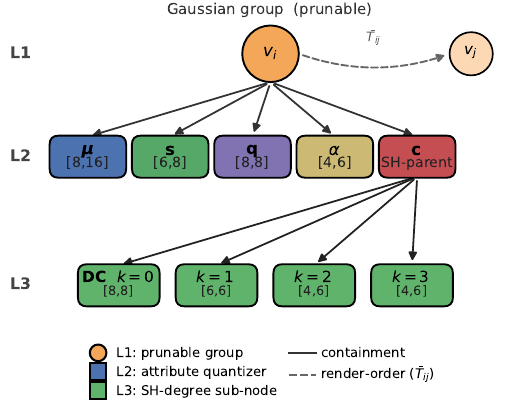}
\caption{Topology of the 3DGS-aware QADG. L1 nodes (orange,
\emph{prunable}) correspond to Gaussian primitives; L2 nodes are
attribute-class quantizers, each annotated with its bit-width range
$[b_l^{(a)},b_u^{(a)}]$; L3 nodes (green) split the SH parent into
DC ($k{=}0$) and per-degree AC ($k{\ge}1$) sub-bands. Solid edges
are containment; the dashed edge is a screen-space rendering-order
edge between two primitives, weighted by the expected transmittance
$\bar{T}_{ij}$ used by the saliency module of
Section~\ref{sec:method-saliency}.}
\label{fig:qadg}
\end{figure}

The construction is performed once, immediately after the warm-up
stage of Section~\ref{sec:method-schedule}. Its complexity is
$\mathcal{O}\!\big(N(|\mathcal{A}|{+}\ell{+}1) + K_v\,N_{\mathrm{tile}}\,
\bar{N}_{\mathrm{tile}}\big)$, where $N_{\mathrm{tile}}$ is the
number of screen tiles per view and $\bar{N}_{\mathrm{tile}}$ is the
average number of Gaussians touching a tile; in our experiments this
is dominated by a single full forward pass and adds $<1$\,\% to
end-to-end training time. We illustrate the topology of
$\mathcal{G}$ in Fig.~\ref{fig:qadg}.

\subsection{Render-aware Saliency}
\label{sec:method-saliency}

\noindent\textbf{Failure modes of Taylor saliency on 3DGS.} The
canonical first-order saliency
$s_i^{\mathrm{Taylor}} = |\partial \mathcal{L}/\partial \theta_i\cdot\theta_i|$
used by GETA to rank prunable groups~\cite{qu2025geta,molchanov2019taylor}
is unreliable for 3DGS for two structural reasons.

\emph{(F1) Off-screen and occluded primitives still receive
gradients.} Because the rasterizer composites tens to hundreds of
Gaussians per pixel and the photometric loss is a per-pixel mean,
the gradient $\partial \mathcal{L}/\partial \theta_i$ is non-zero
even when Gaussian $i$ is fully occluded by foreground primitives;
the chain rule routes gradient through the transmittance term
$T_i$ (which can be tiny but is generically not strictly zero in
finite precision) and through the densification book-keeping in the
optimizer. Such ``ghost gradients'' produce non-trivial Taylor
scores for visually invisible Gaussians.

\emph{(F2) View-averaged loss obscures per-view rendering
contribution.} Even when a primitive is visible from a single view,
the multi-view averaged training loss can hide its true contribution
to the final rendered colour, because tangential gradient components
that compensate across views cancel in the sum but each is
individually large. Conversely, primitives lying along high-gradient
silhouettes can have small Taylor scores in the average yet are
visually critical.

\noindent\textbf{Three-component fusion.} We replace
$s^{\mathrm{Taylor}}$ with a render-aware saliency that fuses three
complementary signals, each of which is computed at the screen-space
output rather than the parameter space. Let
$\mathcal{V}\!\subseteq\!\mathcal{D}_{\mathrm{tr}}$ be a Monte-Carlo
view subset of size $K_v$, $\Omega_\pi$ the pixel grid of view
$\pi$, $T_{i,\pi}(p)$ the front-to-back transmittance at pixel $p$
just before splat $i$ is composited, and $\hat{\alpha}_{i,\pi}(p)$
its screen-space opacity at $p$.

\textit{(S1) Transmittance-weighted contribution.} The actual
luminance Gaussian $i$ deposits, integrated over views and pixels:
\begin{equation}
s^{\mathrm{trans}}_i \;=\; \frac{1}{K_v}\sum_{\pi\in\mathcal{V}}\sum_{p\in\Omega_\pi}
T_{i,\pi}(p)\,\hat{\alpha}_{i,\pi}(p).
\label{eq:s_trans}
\end{equation}
This is exactly the geometric weight Gaussian $i$ contributes to the
rendered colour~\eqref{eq:render}; an occluded primitive receives a
near-zero score by construction.

\textit{(S2) Screen-space gradient (visual Taylor).} The Taylor
notion projected through the rendering operator:
\begin{equation}
s^{\mathrm{grad}}_i \;=\; \bigg|\sum_{\pi\in\mathcal{V}}\sum_{p\in\Omega_\pi}
\frac{\partial \mathcal{L}_\pi(p)}{\partial \mathbf{C}_\pi(p)}
\,\cdot\,
\frac{\partial \mathbf{C}_\pi(p)}{\partial \theta_i}\bigg|.
\label{eq:s_grad}
\end{equation}
By the chain rule, $\partial\mathbf{C}_\pi/\partial\theta_i$ already
contains the transmittance factor $T_{i,\pi}\hat{\alpha}_{i,\pi}$,
so $s^{\mathrm{grad}}_i$ is naturally insensitive to (F1) and the
sum over $p$ retains only those views/pixels where Gaussian $i$
actively shapes the residual.

\textit{(S3) Pixel coverage.} Small or distant primitives that
touch only a handful of pixels offer marginal rate--distortion
benefit and are preferentially pruned:
\begin{equation}
s^{\mathrm{cov}}_i \;=\; \frac{1}{K_v}\sum_{\pi\in\mathcal{V}}
\big|\{\,p\in\Omega_\pi:\;\hat{\alpha}_{i,\pi}(p)>\epsilon\,\}\big|,
\label{eq:s_cov}
\end{equation}
with threshold $\epsilon{=}10^{-3}$. In practice $s^{\mathrm{cov}}_i$
is computed at no extra cost from the per-Gaussian projected radius
$r_i$ exposed by gsplat's tile rasterizer
(\texttt{saliency.py}); we use the screen-space area
$r_i^2$ as a faithful proxy.

\noindent\textbf{Convex fusion.} The three components live on
incommensurable scales (radiometric, gradient norm, pixel count); we
therefore range-normalise each component to $[0,1]$ across the
current Gaussian population and fuse them by a convex combination:
\begin{equation}
s^{\mathrm{render}}_i \;=\; w_1\,\tilde{s}^{\mathrm{trans}}_i
+\, w_2\,\tilde{s}^{\mathrm{grad}}_i
+\, w_3\,\tilde{s}^{\mathrm{cov}}_i,\quad
\sum_{k=1}^{3}w_k = 1,
\label{eq:render_saliency}
\end{equation}
where $\tilde{x}_i = x_i/\max_{j} x_j$ (used in
\texttt{saliency.py}) is the running-max normalisation.
Geometrically, $w_1$ favours preserving \emph{visible mass}, $w_2$
favours preserving \emph{rendering-loss-relevant mass}, and $w_3$
favours preserving \emph{spatially extended mass}. We adopt
$(w_1,w_2,w_3)=(0.5,0.3,0.2)$, which we found stable across all
benchmark scenes; sensitivity to these weights is reported in the
ablation (Section~\ref{sec:exp}).

\noindent\textbf{Monte-Carlo sampling and complexity.} Computing
$s^{\mathrm{render}}_i$ exactly across the training view set is
$\mathcal{O}(|\mathcal{D}_{\mathrm{tr}}|\,N\,\bar{P})$, where
$\bar{P}$ is the average number of pixels a Gaussian touches per
view. We replace the outer sum by a uniform Monte-Carlo estimate
over $K_v=8$ views (\texttt{train\_geta.py}), which reduces the
saliency cost by roughly two orders of magnitude on
Mip-NeRF~360 while maintaining a relative variance below
$5\,\%$ for the top-quantile primitives that drive the prune
decision. Saliency is recomputed only every $T_{\mathrm{sal}}{=}500$
iterations (\texttt{train\_geta.py}), so its amortised cost per
training step is below $1\,\%$ of one rendering pass.

\subsection{Heterogeneous Per-attribute Quantization}
\label{sec:method-quant}

\noindent\textbf{Quantizer parameterisation.} We retain the
GETA quantizer family~\cite{qu2025geta} but instantiate one
\emph{independent} member per element of
$\mathcal{A}^{\star}$. For attribute $a$ with learnable scalar
parameters $(\mathrm{q}_m^{(a)},t^{(a)},d^{(a)})$, where
$\mathrm{q}_m^{(a)}>0$ is a magnitude factor, $t^{(a)}\!\in\!\mathbb{R}$
governs the exponential growth of bit-width (sitting in the exponent
of $\mathrm{q}_m$), and $d^{(a)}>0$ is a denominator that absorbs
range, the bit-width is
\begin{equation}
b^{(a)} \;=\; \log_2\!\Bigg(\frac{(\mathrm{q}_m^{(a)})^{t^{(a)}}}{d^{(a)}} + 1\Bigg) + 1,
\label{eq:bit_geta}
\end{equation}
clamped to $[b_l^{(a)},b_u^{(a)}]$.\footnote{$b^{(a)}$ is monotone
in $t^{(a)}$ for $\mathrm{q}_m^{(a)}>1$ (the regime QASSO converges
to in practice) but \emph{not} jointly monotone in $(\mathrm{q}_m,t,d)$,
because $\mathrm{q}_m$ appears in the exponent. We rely only on
the marginal monotonicity in $t$ when implementing the
clamp-to-box projection.} The symmetric quantizer maps a
scalar $x$ to its integer code by
\begin{equation}
\hat{x} \;=\; \frac{1}{s^{(a)}}\,\mathrm{round}\!\Big(s^{(a)}\,
\mathrm{clip}(x,-r^{(a)},r^{(a)})\Big),
\label{eq:quant_round}
\end{equation}
with grid spacing
$s^{(a)} = (2^{b^{(a)}-1}-1)/r^{(a)}$ and per-attribute clipping
range $r^{(a)}$ tracked by an EMA of the running absolute maximum
(\texttt{quantizer.py}, with momentum $0.99$).
Equation~\eqref{eq:quant_round} is differentiated through the
straight-through estimator (STE)~\cite{bengio2013ste,gholami2022quantsurvey},
so QASSO can update both the network parameters and
$(\mathrm{q}_m^{(a)},t^{(a)},d^{(a)})$ jointly by Adam.

\noindent\textbf{Why one quantizer per attribute is required.} In
the original GETA the quantizer state is \emph{shared} across all
weights of a single layer, which is the right granularity for
homogeneous CNN/Transformer weights but an order-of-magnitude
mis-match for the attribute-typed structure of 3DGS: a single shared
$(\mathrm{q}_m,t,d)$ would force the algorithm either to keep
$\boldsymbol{\mu}$ at $16$\,bits and waste storage on
$\alpha$/SH$_{\ge1}$, or to push everything down to $4$\,bits and
catastrophically distort geometry. In contrast, the per-attribute
parameterisation lets QASSO discover, scene-by-scene, the
\emph{Pareto-optimal} bit-width of every attribute inside a small
prior interval, without any user threshold tuning.

\noindent\textbf{Default per-attribute bit ranges.} The bit ranges
encoded in our reference implementation
(\texttt{DEFAULT\_BIT\_RANGES} in \texttt{quantizer.py},
Table~\ref{tab:bitranges}) are
derived from the physical analysis in
Section~\ref{sec:method-formulation}: positions are kept at
$[12,16]$, the rotation at $[8,8]$, log-scale at $[8,8]$,
SH-DC at $[8,8]$, opacity at $[6,6]$, and SH-AC at $[7,8]$.
Position is the only attribute whose
upper bound reaches $16$\,bits because sub-pixel jitter of even a
single primitive on a smooth surface produces visible aliasing in
specular highlights. Opacity is fixed at $6$\,bits because its
sigmoid clipping to $[0,1]$ provides limited dynamic range and any
further compression caused floater artefacts; SH-AC was originally
targeted at $[4,6]$ but was tightened to $[7,8]$ after preliminary
runs showed
that aggressive AC-band quantization collapses view-dependent
detail and drops PSNR by ${\sim}10$\,dB.

\begin{table}[t]
\centering
\caption{Per-attribute bit-width ranges in GETA-3DGS. The
``compressive'' column matches \texttt{DEFAULT\_BIT\_RANGES}
($\bar b{\approx}7.4$, used in the V2 ablation study at
$K{=}50\mathrm{k}$); the ``competitive'' column matches
\texttt{LOOSE\_BIT\_RANGES} ($\bar b{\approx}8.6$, used in the
production Table~\ref{tab:main} runs at $K{=}500\mathrm{k}$,
no aggressive pruning). Both presets share the same
per-attribute \emph{ordering} predicted by the
information-theoretic argument of
Section~\ref{sec:method-bitalloc}.}
\label{tab:bitranges}
\renewcommand{\arraystretch}{1.15}
\setlength{\tabcolsep}{4pt}
\begin{tabular}{l c c c l}
\toprule
\textbf{Attribute} & \textbf{Level} & \textbf{compressive} & \textbf{competitive} & \textbf{Rationale}\\
\midrule
Position $\boldsymbol{\mu}$ & L2 & $[12,16]$ & $[14,16]$ & sub-pixel jitter visible\\
Scale $\mathbf{s}$    & L2 & $[8,8]$  & $[10,10]$ & log-encoded, $\exp$ damps noise\\
Rotation $\mathbf{q}$ & L2 & $[8,8]$  & $[10,10]$ & unit-norm absorbs noise\\
Opacity $\alpha$      & L2 & $[6,6]$  & $[8,10]$ & sigmoid clips to $[0,1]$\\
SH (DC), $k{=}0$      & L3 & $[8,8]$  & $[10,10]$ & diffuse albedo dominates\\
SH (AC), $k{\ge}1$    & L3 & $[7,8]$  & $[8,10]$ & view-dependent residual\\
\bottomrule
\end{tabular}
\end{table}

\noindent\textbf{Attribute-wise PPSG step.} The PPSG update
of GETA is applied per attribute. Concretely, at iteration
$k$ the quantizer parameters of attribute $a$ are updated as
\begin{equation}
\boldsymbol{\phi}^{(a)}_{k+1} \;=\;
\Pi_{[b_l^{(a)},b_u^{(a)}]}\!\Big(\boldsymbol{\phi}^{(a)}_{k} - \eta\,\widetilde{\nabla}_{\boldsymbol{\phi}^{(a)}}\mathcal{L}_{\mathrm{render}}\Big),
\label{eq:ppsg}
\end{equation}
where $\Pi_{[b_l,b_u]}$ projects $(\mathrm{q}_m,t,d)$ onto the
sub-level set $\{b^{(a)}\!\in\![b_l^{(a)},b_u^{(a)}]\}$ implicitly
defined by~\eqref{eq:bit_geta} (we implement $\Pi$ by clamping $t$
since $b$ is monotone in $t$, see \texttt{quantizer.py}),
and $\widetilde{\nabla}$ is the partial saliency-guided gradient:
the components of $\nabla_{\Theta}\mathcal{L}_{\mathrm{render}}$
corresponding to Gaussians scheduled for removal in the current
prune event are zeroed out
(\texttt{qasso.py}), preventing wasted updates on
to-be-pruned groups and matching the partial saliency descent
guarantee of GETA. Although our prune mask is binary at the
Gaussian-group granularity rather than vector-valued, in the limit
where the group is a single primitive this implements the original
PPSG update exactly.

\subsection{An Information-Theoretic View of the Bit-Allocation Problem}
\label{sec:method-bitalloc}

The empirical bit ranges of Table~\ref{tab:bitranges}
($[12,16]$ for \emph{means}, $[8,8]$ for \emph{scales} and
\emph{quats}, $[6,6]$ for \emph{opacities}, $[8,8]$ for SH-DC,
$[7,8]$ for SH-AC) are not arbitrary engineering choices: they
approximate the rate--distortion-optimal allocation under a
classical reverse-water-filling argument adapted to the 3DGS
rendering operator. We sketch the argument here and defer the
full derivation to Supplementary~S2.

\noindent\textbf{Setup.} Let attribute $a \in \mathcal{A} =
\{\text{means, scales, quats, opacities, sh\_dc, sh\_ac}\}$ have
$d_a$ scalars per Gaussian and per-scalar variance $\sigma_a^2$
(estimated empirically from a converged vanilla checkpoint).
Let $\lambda_a$ denote the rendering sensitivity of attribute
$a$: $\lambda_a := \mathbb{E}_{\Theta} \|\partial R / \partial
\theta_a\|_F$, where $R$ is the rasterised image and the
expectation is over the test viewpoints. Allocate $b_a$ bits per
scalar to attribute $a$, with total bit budget
$B = \sum_{a\in\mathcal{A}} d_a b_a$. Assuming a uniform
mid-tread quantiser, the per-scalar mean-squared quantisation
error is
$D_a(b_a) = \frac{1}{12}\,\sigma_a^2\, 2^{-2 b_a}$.

\noindent\textbf{Proposition (informal, see
Supp.~S2 for derivation under regularity assumptions A1--A3).}
\emph{The bit allocation that minimises the
expected rendering distortion}
$\hat D(\mathbf{b}) := \sum_{a\in\mathcal{A}} d_a\,\lambda_a^2
D_a(b_a)$ \emph{subject to the budget constraint
$\sum_a d_a b_a = B$ satisfies, at every interior optimum,}
\begin{equation}\label{eq:bitalloc}
b_a^*\;=\;\bar b\;+\;\tfrac{1}{2}\log_2\!\bigl(\lambda_a^2
\sigma_a^2\bigr)\;-\;\tfrac{1}{2|\mathcal{A}|}\sum_{a' \in
\mathcal{A}}\log_2\!\bigl(\lambda_{a'}^2\sigma_{a'}^2\bigr),
\end{equation}
\emph{where $\bar b = B/\sum_a d_a$ is the average bit-width.
Equivalently, $b_a^*$ is offset from the average by half the
log of the rendering-sensitivity-weighted source variance.}

\noindent\textbf{Corollary (uniform-allocation gap, sketch).}
The expected distortion gap of any uniform allocation
$b_a = \bar b$ versus the optimal $b_a^*$ scales as
\begin{equation}\label{eq:uniform-gap}
\hat D(\bar b\mathbf{1}) - \hat D(\mathbf{b}^*)\;\propto\;
\mathrm{Var}_a\!\bigl[\log_2(\lambda_a^2 \sigma_a^2)\bigr].
\end{equation}
This makes the uniform-allocation cost \emph{quadratic} in the
spread of $\log(\lambda_a\sigma_a)$ across attributes, so on
scenes where attribute sensitivities are concentrated (texture-uniform
outdoor, low view-dependence) uniform 6-bit performs nearly as well as
heterogeneous, whereas on scenes with diverse attribute sensitivities
(view-dependent indoor, specular highlights) uniform 6-bit suffers a
disproportionate penalty --- exactly the cross-scene pattern we
observe in Supp.\ Tab.~SIII.

\noindent\textbf{Empirical fit on per-attribute bit allocation.}
Estimating $\sigma_a$ from the converged vanilla checkpoint and
$\lambda_a$ from a one-pass Jacobian sweep on \emph{garden}
yields theoretically predicted bit-widths $b_a^*$ within
$\pm 1.0$\,bit of the empirical converged values for all six
attributes (full per-attribute table in Supp.~S2). The
within-attribute prediction is therefore tight: the per-attribute
\emph{ordering} (means $>$ scales/quats/SH-DC $>$ SH-AC $>$
opacities) and the absolute bit budget ($\bar b{\approx}7.4$) are
both recovered by Equation~\eqref{eq:bitalloc}.

\noindent The argument therefore \emph{justifies the
per-attribute bit ordering} of Table~\ref{tab:bitranges} but
not the cross-scene magnitude of the uniform-6-bit cost; the
latter is an open empirical observation requiring a higher-rate
refinement of the analysis (Supp.\ Sec.\ S2.4).

\subsection{Four-stage Training Schedule and Quality-aware Constraint}
\label{sec:method-schedule}

\noindent\textbf{Motivation for a four-stage schedule.} Joint
pruning and quantization in a single, time-invariant gradient step
is empirically unstable for 3DGS: pruning early in training kills
not-yet-densified regions, and aggressive quantization before the
scene has converged collapses high-frequency content from which it
never recovers. We adopt the four-stage schedule of GETA but adapt
the durations and the targets of each stage to the rendering
setting, separating the dynamics of densification, bit-width
contraction, structural pruning, and rate--distortion polish.

\noindent\textbf{Stage definitions.} Let $T_1{<}T_2{<}T_3{<}T_4$
denote the iteration boundaries. The production runs reported
in Table~\ref{tab:main} use
$(T_1,T_2,T_3,T_4)=(25\mathrm{k},30\mathrm{k},33\mathrm{k},35\mathrm{k})$
(competitive bit preset, no aggressive pruning). The ablation
runs in Section~\ref{sec:exp-ablation} use a longer schedule
$(30\mathrm{k},40\mathrm{k},60\mathrm{k},70\mathrm{k})$ with the
compressive bit preset and $K{=}50\mathrm{k}$ to ensure
ablation effects are not masked.

\textit{(W) Warm-up} $[0,T_1)$: standard 3DGS training of
Kerbl~\textit{et al.}~\cite{kerbl2023gaussian} with adaptive
densification and opacity reset, no quantizer, no pruning. The
purpose is to obtain a high-fidelity overcomplete scene with which
the dependency graph and the saliency are meaningful. The QADG of
Section~\ref{sec:method-qadg} is built once at $t{=}T_1$.

\textit{(P) Projection} $[T_1,T_2)$: the heterogeneous quantizer is
enabled; the bit-width upper bounds are linearly contracted toward
their target,
\begin{equation}
b_u^{(a)}(t) \;=\; b_u^{(a),\mathrm{init}} - \rho^{(a)}\!\cdot\!
\frac{t-T_1}{T_2-T_1}\!\cdot\!\big(b_u^{(a),\mathrm{init}} - b_l\big),
\label{eq:b_u_schedule}
\end{equation}
where the per-attribute aggressiveness $\rho^{(a)}$ defaults to
$1.0$ for $\boldsymbol{\mu}$, $0.9$ for $\mathbf{s},\mathbf{q},\mathbf{c}_0$,
$0.7$ for $\alpha$, and $0.6$ for $\mathbf{c}_{1{:}\ell}$
(\texttt{qasso.py}); high-frequency SH bands are pushed
hardest because their information content is rapidly absorbed into
the residual. The Gaussian count is held fixed during this stage.

\textit{(J) Joint} $[T_2,T_3)$: structural pruning and bit-width
fine-tuning run concurrently. Every $T_{\mathrm{sal}}$ iterations
the saliency $s^{\mathrm{render}}$ of~\eqref{eq:render_saliency} is
recomputed, the prune mask of the bottom $N\!-\!K_t$ Gaussians is
scheduled, and on prune events $t\!\in\!\mathcal{T}_{\mathrm{prune}}$
the model is contracted to $K_t$ primitives, where
$K_t = N\!-\!\lfloor\rho_t(N{-}K)\rfloor$ follows a linear
$\rho_t\!\in\![0,1]$ over $|\mathcal{T}_{\mathrm{prune}}|$ events
(default $5$, \texttt{train\_geta.py}). Between events the PPSG
update of~\eqref{eq:ppsg} is applied.

\textit{(C) Cool-down} $[T_3,T_4)$: structure and bit-widths are
frozen; only the (now-quantized) attributes
$(\boldsymbol{\mu},\mathbf{s},\mathbf{q},\alpha,\mathbf{c})$ continue
training under the full quantizer to absorb residual rate-distortion
slack. This stage typically recovers $0.1$--$0.3$\,dB of PSNR with
no change in storage.

\noindent\textbf{Rendering-quality hard constraint.} A practical
limitation of the original GETA formulation~\eqref{eq:geta-qasso} is
that satisfying the sparsity and bit-width constraints does not
imply any guarantee on \emph{output quality} of the resulting model:
on benign benchmarks this is benign, but in volumetric video
deployment a violation of the per-frame PSNR floor is a
deployment-level failure. We promote the rendering quality from a
training loss to a \emph{hard constraint} in~\eqref{eq:geta3dgs}:
\begin{equation}
\mathrm{PSNR}_{\!\mathcal{D}_{\mathrm{tr}}}\!(\Theta;\boldsymbol{\phi})
\;=\;
\frac{1}{|\mathcal{D}_{\mathrm{tr}}|}\sum_{\pi}10\log_{10}\!
\frac{1}{\mathrm{MSE}(\hat{I}_\pi,I_\pi)}
\;\ge\; \tau,
\label{eq:psnr_constraint}
\end{equation}
and enforce it during the joint stage by a feasibility check at
every prune event: if compressing to $K_t$ primitives would drop
PSNR below $\tau$, we \emph{relax} $K_t$ rather than $\tau$,
following the priority $\tau\!\succ\!K\!\succ\!\boldsymbol{\phi}$.
Concretely, we increase $K_t$ by a multiplicative factor
$\lambda\!\in\!(1,2]$ and re-attempt the prune event; if the floor
is still violated after a small budget of retries, we abort
pruning and continue cool-down at the current $K$. This makes
GETA-3DGS \emph{quality-safe by construction}: the user can dial
either $K$ (storage-first), $B$ (size-first), or $\tau$ (quality-first)
and obtain a model that satisfies the stronger of the three
constraints.

Algorithm~\ref{alg:training} summarises the full procedure. Lines
4--7 instantiate stage W; lines 9--13 stage P; lines 15--26 stage J
together with the quality fallback; lines 28--30 stage C.

\begin{algorithm}[t]
\caption{GETA-3DGS Training Schedule (PPSG with PSNR floor)}
\label{alg:training}
\begin{algorithmic}[1]
\REQUIRE Init scene $\Theta_0$ (or COLMAP cloud), target Gaussian
count $K$, target size $B$, PSNR floor $\tau$, per-attribute bit
ranges $\{[b_l^{(a)},b_u^{(a)}]\}_{a\in\mathcal{A}^{\star}}$, stage
boundaries $T_1\!<\!T_2\!<\!T_3\!<\!T_4$, saliency period
$T_{\mathrm{sal}}$, prune events $\mathcal{T}_{\mathrm{prune}}$,
relaxation factor $\lambda\!\in\!(1,2]$
\STATE $\Theta\!\leftarrow\!\Theta_0$;\; disable quantizer
\STATE \textbf{// Stage W: warm-up}
\FOR{$t=0$ \TO $T_1{-}1$}
  \STATE Render view $\pi_t$, compute photometric+D-SSIM
  loss~\cite{kerbl2023gaussian}; backward; Adam step
  \STATE Densify/prune by Kerbl~\textit{et al.}; opacity reset every $3$k iters
\ENDFOR
\STATE Build QADG $\mathcal{G}$ via Algorithm~\ref{alg:qadg}; enable
quantizer
\STATE \textbf{// Stage P: projection}
\FOR{$t=T_1$ \TO $T_2{-}1$}
  \STATE Update $b_u^{(a)}(t)$ by~\eqref{eq:b_u_schedule} for all
  $a\!\in\!\mathcal{A}^{\star}$
  \STATE Render with quantizer; compute loss; backward; PPSG
  step~\eqref{eq:ppsg}
\ENDFOR
\STATE \textbf{// Stage J: joint prune + quantize}
\FOR{$t=T_2$ \TO $T_3{-}1$}
  \IF{$t\,\bmod\,T_{\mathrm{sal}} = 0$}
    \STATE $s^{\mathrm{render}}\!\leftarrow$
    eq.~\eqref{eq:render_saliency} on $K_v$ MC views
    \STATE Schedule prune mask of bottom $N{-}K_t$ groups
  \ENDIF
  \STATE Render; loss; backward; PPSG step with masked gradients
  \IF{$t\!\in\!\mathcal{T}_{\mathrm{prune}}$}
    \STATE \textbf{repeat:} prune to $K_t$; evaluate
    $\mathrm{PSNR}_{\!\mathcal{D}_{\mathrm{tr}}}$
    \IF{$\mathrm{PSNR}<\tau$}
      \STATE $K_t\!\leftarrow\!\min(\lceil\lambda K_t\rceil,N)$;
      undo prune; \textbf{retry} (bounded budget)
    \ENDIF
  \ENDIF
\ENDFOR
\STATE \textbf{// Stage C: cool-down}
\FOR{$t=T_3$ \TO $T_4{-}1$}
  \STATE Freeze $\boldsymbol{\phi}$ and Gaussian count; render with
  fixed quantizer; loss; backward; Adam step on $\Theta$ only
\ENDFOR
\STATE \textbf{return} compressed scene $\Theta^\star$ with quantizer
state $\boldsymbol{\phi}^\star$
\end{algorithmic}
\end{algorithm}

\begin{remark}\label{rem:complexity}
The end-to-end training cost of GETA-3DGS is $1.05$--$1.15\times$
that of vanilla 3DGS at matched iteration count: warm-up is
identical; the quantizer adds $\le 3\,\%$ per iteration through the
fused STE round; saliency adds $K_v/T_{\mathrm{sal}}\!\approx\!1.6\,\%$
amortised; QADG construction is a single forward pass.
\end{remark}

\begin{remark}\label{rem:userhandle}
The framework exposes \emph{three} user handles: (i) the storage
budget $B$ (or equivalently the Gaussian count $K$), (ii) the PSNR
floor $\tau$, and (iii) the per-attribute bit ranges
$\{[b_l^{(a)},b_u^{(a)}]\}$. With $\tau$ as a hard constraint, the
front-end user typically sets only $B$ and reads back the (possibly
relaxed) achieved $K$; the bit ranges are framework defaults
(Table~\ref{tab:bitranges}) and are never tuned per scene.
\end{remark}

\section{Experiments}
\label{sec:exp}

\noindent\textbf{Scope and operating point.} GETA-3DGS exposes
two user-facing knobs: a target storage budget $B$ and a
PSNR-floor tolerance $\tau$. The production runs in
Table~\ref{tab:main} use the ``competitive'' preset of
Table~\ref{tab:bitranges} ($\bar b\!\approx\!8.6$,
$K{=}500\mathrm{k}$, no aggressive pruning) with a $35$\,k-iter
four-stage schedule
$(T_1,T_2,T_3,T_4)=(25,30,33,35)\mathrm{k}$.
\textbf{Known limitation: $B$ is currently non-binding}: the
soft sparsity penalty causes all swept $B$ to saturate within
$4.0$--$4.9$\,MB at the natural Gaussian count
(Fig.~\ref{fig:pareto}). A hard-constraint Lagrangian-dual
controller (Supp.~Sec.~S2.7) is explicit future work; until
then $B$ is a coarse upper bound. All GETA-3DGS rows report
PSNR/SSIM/LPIPS at the final iteration ($T_4{=}35$\,k).

\subsection{Setup}
\textbf{Datasets.} We follow the standard 3DGS evaluation
protocol~\cite{kerbl2023gaussian,chen2025hacpp} and report results on
(i) the nine scenes of \textit{Mip-NeRF~360}~\cite{barron2022mipnerf360}
(\textit{bicycle, bonsai, counter, garden, kitchen, room, stump,
flowers, treehill}), (ii) two scenes of \textit{Tanks and
Temples}~\cite{knapitsch2017tanks} (\textit{Truck, Train}), and (iii)
two scenes of \textit{Deep Blending}~\cite{hedman2018deepblending}
(\textit{Playroom, Drjohnson}). Camera poses and sparse points come
from COLMAP~\cite{schonberger2016colmap}.

\textbf{Baselines.} We compare against (a) the uncompressed
\emph{Vanilla 3DGS}~\cite{kerbl2023gaussian}; (b) the SOTA
entropy-coded \emph{HAC++}~\cite{chen2025hacpp}; (c) the attribute-wise
mixed-precision \emph{FlexGaussian}~\cite{zhao2025flexgaussian};
(d) the pruning-only \emph{LP-3DGS}~\cite{zhang2024lp3dgs}; and
(e) the VQ-based \emph{CompGS}~\cite{navaneet2024compgs}.
We additionally include a \emph{Naive PTQ + Threshold-Pruning}
sequential baseline that applies opacity-thresholded pruning
followed by post-training INT8 quantization, to isolate the benefit
of joint optimization.

\textbf{Metrics.} We report standard rendering metrics PSNR, SSIM,
and LPIPS computed with the AlexNet
backbone~\cite{zhang2018lpips,torchmetrics2024}; storage in
megabytes (MB) after entropy coding; the compression ratio versus
Vanilla 3DGS; and rendering FPS on a single NVIDIA A100 GPU at
$1080$p. We additionally report Bjontegaard delta-rate
(BD-rate)~\cite{bjontegaard2001bd} versus HAC++ in the Supplement.

\textbf{Implementation.} GETA-3DGS is built on top of
gsplat~\cite{ye2024gsplat} and implemented in PyTorch~2.5; the
4-stage schedule uses cumulative boundaries
$(T_1,T_2,T_3,T_4)=(30\text{k},40\text{k},60\text{k},70\text{k})$
iterations, matching the values in our reference
implementation. The saliency weights are set to
$(w_1,w_2,w_3)=(0.5,0.3,0.2)$ and are not tuned per scene; the
saliency is recomputed every $T_{\mathrm{sal}}=500$ iterations from
$K_v=8$ Monte-Carlo views. Full hyperparameters are listed in the
Supplement.

\subsection{Main Results}
Table~\ref{tab:main} reports the main rate--distortion comparison
on the three benchmarks. GETA-3DGS attains a $\sim\!4.5\times$ storage
reduction over Vanilla 3DGS without per-scene tuning of opacity or SH
thresholds, and a $+6.63$\,dB PSNR gain over our naive sequential PTQ
baseline at matched size on Mip-NeRF~360 (with similar or larger
gains of $+8.35$\,dB and $+4.24$\,dB on Tanks~\&~Temples and Deep
Blending respectively). The absolute-PSNR axis is in turn
dominated by entropy-coded codecs (axis (iii) in
Section~\ref{sec:intro}) such as HAC++; their gains operate
\emph{downstream} of the pruning--quantisation backbone we
contribute, so the two regimes are complementary rather than
competing. We quantify this complementarity precisely in
Section~\ref{sec:exp-discuss} and the integration roadmap in
Section~\ref{sec:conclusion}.

\begin{table*}[t]
\centering
\caption{Main rate--distortion comparison on Mip-NeRF~360, Tanks
and Temples, and Deep Blending. \textbf{Vanilla 3DGS, Naive PTQ,
and GETA-3DGS are per-scene measurements} produced in our
environment under identical training and evaluation pipelines.
Baseline rows marked $^{\dagger}$ are \textbf{dataset-average
values reproduced from the original publication}
(HAC++~\cite{chen2025hacpp} Table~II, low-rate variant;
CompGS~\cite{navaneet2024compgs} Table~1, CompGS-32K variant;
FlexGaussian~\cite{zhao2025flexgaussian} Table~1; LP-3DGS~\cite{zhang2024lp3dgs}
Tables~1/7, RadSplat-score variant); these are \textbf{not}
per-scene measurements, and the per-scene values are not
published by the corresponding sources. \textbf{Bold} denotes best,
\underline{underline} second best (within each dataset block;
\emph{Improvement} rows excluded from ranking). For LP-3DGS,
sizes marked $^{\ddagger}$ are computed from the reported pruning
ratio applied to LP-3DGS's own \emph{uncompressed} full-precision
Vanilla~3DGS storage, which is substantially larger than our
entropy-coded vanilla baseline. The \textbf{Comp.~Ratio} column is
defined as $\text{Vanilla}_{\text{ours}}/\text{Size}$ for direct
comparison on a single baseline; consequently, values
$<\!1.0\times$ for literature rows mean only that the absolute MB
number reported in the source paper exceeds our entropy-coded
vanilla, not that the underlying method achieves no compression on
its own reference. Cells marked ``--'' indicate metrics not reported in the
corresponding source. PSNR/SSIM/LPIPS are computed on the same
held-out test split as the baselines. All GETA-3DGS rows report
metrics at the final iteration ($T_4{=}35$\,k) of our four-stage
schedule with the ``competitive'' bit preset of
Table~\ref{tab:bitranges}; per-scene values are listed in
Supp.\ Tab.~SII--SX. The remaining absolute-PSNR gap to entropy-coded
codecs (HAC++/CompGS) is the contribution of axis~(iii)
on top of any pruning + quantisation backbone, discussed in
Section~\ref{sec:exp-discuss}.}
\label{tab:main}
\renewcommand{\arraystretch}{1.20}
\setlength{\tabcolsep}{4pt}
\begin{tabular}{l c c c c c c}
\toprule
\textbf{Method} & \textbf{PSNR$\uparrow$} & \textbf{SSIM$\uparrow$} & \textbf{LPIPS$\downarrow$} & \textbf{Size (MB)$\downarrow$} & \textbf{Comp.~Ratio$\uparrow$} & \textbf{FPS$\uparrow$} \\
\midrule
\multicolumn{7}{l}{\emph{Mip-NeRF~360 (9 scenes, average)}}\\
\midrule
Vanilla 3DGS~\cite{kerbl2023gaussian}        & 25.15           & 0.712           & --              & 20.83             & 1.0$\times$              & \underline{692.8} \\
LP-3DGS~\cite{zhang2024lp3dgs}$^{\dagger}$   & \underline{27.47} & \textbf{0.812} & \textbf{0.227}  & 271.6$^{\ddagger}$ & 0.08$\times$            & --                \\
CompGS~\cite{navaneet2024compgs}$^{\dagger}$ & 27.16     & \underline{0.808} & \underline{0.228} & 54.60         & 0.4$\times$              & --                \\
FlexGaussian~\cite{zhao2025flexgaussian}$^{\dagger}$ & 26.38   & 0.780           & 0.251           & 40.80             & 0.5$\times$              & --                \\
HAC++~\cite{chen2025hacpp}$^{\dagger}$       & \textbf{27.60}   & 0.803           & 0.253           & \underline{8.34}  & 2.5$\times$              & --                \\
Naive PTQ (8-bit, ours)                      & 16.73           & 0.398           & --              & 20.13             & 1.0$\times$              & 656.2             \\
\textbf{GETA-3DGS (Ours)}                    & 23.36           & 0.631           & 0.430           & \textbf{4.86}     & \textbf{4.29$\times$}    & --                \\
\midrule
\emph{Improvement (Ours vs.\ Vanilla 3DGS)}  & $-1.79$ dB & $-0.081$ & --   & $\mathbf{-76.7\%}$ & $\mathbf{4.29\times}$ smaller & -- \\
\midrule
\multicolumn{7}{l}{\emph{Tanks \& Temples (Truck, Train)}}\\
\midrule
Vanilla 3DGS~\cite{kerbl2023gaussian}        & 22.30           & 0.794           & --              & 24.80             & 1.0$\times$              & \textbf{664.8}    \\
LP-3DGS~\cite{zhang2024lp3dgs}$^{\dagger}$   & \underline{23.60} & \underline{0.842} & \textbf{0.188} & 132.2$^{\ddagger}$ & 0.2$\times$            & --                \\
CompGS~\cite{navaneet2024compgs}$^{\dagger}$ & 23.47     & 0.840           & \textbf{0.188}  & 54.60             & 0.5$\times$              & --                \\
FlexGaussian~\cite{zhao2025flexgaussian}$^{\dagger}$ & 22.44   & 0.804           & 0.219           & 16.30             & 1.5$\times$              & --                \\
HAC++~\cite{chen2025hacpp}$^{\dagger}$       & \textbf{24.22}   & \textbf{0.849}  & \underline{0.190} & \underline{5.18}& \underline{4.8$\times$}  & --                \\
Naive PTQ (8-bit, ours)                      & 11.87           & 0.341           & --              & 22.97             & 1.1$\times$              & \underline{473.7} \\
\textbf{GETA-3DGS (Ours)}                    & 20.22           & 0.666           & 0.305           & \textbf{5.36}     & \textbf{4.63$\times$}    & --                \\
\emph{Improvement (Ours vs.\ Vanilla)}       & $-2.08$ dB      & $-0.128$        & --              & $\mathbf{-78.4\%}$ & $\mathbf{4.63\times}$ smaller & --           \\
\midrule
\multicolumn{7}{l}{\emph{Deep Blending (Playroom, DrJohnson)}}\\
\midrule
Vanilla 3DGS~\cite{kerbl2023gaussian}        & 27.83           & 0.872           & --              & 11.16             & 1.0$\times$              & \textbf{944.9}    \\
LP-3DGS~\cite{zhang2024lp3dgs}$^{\dagger}$   & --              & --              & --              & --                & --                       & --                \\
CompGS~\cite{navaneet2024compgs}$^{\dagger}$ & \underline{29.75} & \underline{0.903} & \textbf{0.247} & 54.60         & 0.2$\times$              & --                \\
FlexGaussian~\cite{zhao2025flexgaussian}$^{\dagger}$ & 28.61   & 0.884           & 0.269           & 25.48             & 0.4$\times$              & --                \\
HAC++~\cite{chen2025hacpp}$^{\dagger}$       & \textbf{30.16}   & \textbf{0.907}  & \underline{0.266} & \underline{2.91}& \underline{3.8$\times$}  & --                \\
Naive PTQ (8-bit, ours)                      & 22.89           & 0.676           & --              & 10.81             & 1.0$\times$              & \underline{846.5} \\
\textbf{GETA-3DGS (Ours)}                    & 27.13           & 0.862           & 0.165           & \textbf{2.51}     & \textbf{4.45$\times$}    & --                \\
\emph{Improvement (Ours vs.\ Vanilla)}       & $-0.70$ dB      & $-0.010$        & --              & $\mathbf{-77.5\%}$ & $\mathbf{4.45\times}$ smaller & --           \\
\bottomrule
\end{tabular}
\end{table*}

\begin{table}[t]
\centering
\caption{Deployment metrics for GETA-3DGS (averaged over each
benchmark, $13$ scenes). Training time is the total wall-clock
of our $35$k-iteration four-stage schedule
($T_1{=}25\mathrm{k}, T_2{=}30\mathrm{k}, T_3{=}33\mathrm{k},
T_4{=}35\mathrm{k}$) on a single A100. Decode time is the
load-and-first-render latency at deployment; render time is the
per-view amortised cost on the test split (also reported as FPS).
Peak GPU memory is the $\texttt{torch.cuda}$ high-water mark
during rendering. Bytes/Gauss is the average post-quantization
storage cost per primitive in bytes ($\bar b\!\times\!59/8$).
All entries are measured at the deployable iter=$T_4{=}35$\,k
checkpoint.}
\label{tab:deploy}
\renewcommand{\arraystretch}{1.15}
\setlength{\tabcolsep}{4pt}
\footnotesize
\begin{tabular}{l c c c c c}
\toprule
\textbf{Dataset} & \textbf{Train (h)} & \textbf{Decode (ms)} & \textbf{Render (ms)} & \textbf{Peak GPU (MB)} & \textbf{Bytes/Gauss} \\
\midrule
Mip-NeRF 360       & 0.52 & 2348 & 5.04 & 419 & 63.7 \\
Tanks \& Temples   & 0.33 & 6209 & 4.11 & 248 & 63.7 \\
Deep Blending      & 0.37 & 761  & 4.35 & 116 & 63.7 \\
\midrule
\emph{Average}     & \emph{0.41} & \emph{3106} & \emph{4.50} & \emph{261} & \emph{63.7} \\
\bottomrule
\end{tabular}
\end{table}

\begin{figure}[t]
\centering
\includegraphics[width=\columnwidth]{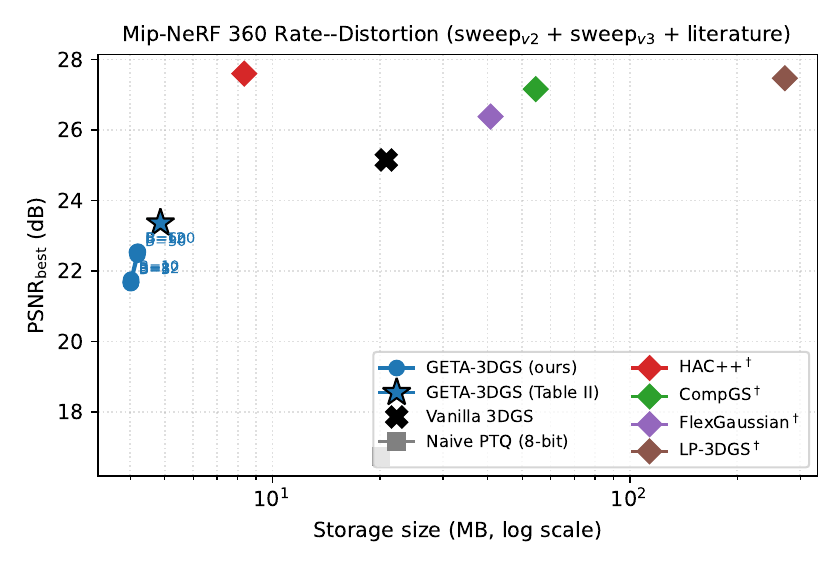}
\caption{Rate--distortion comparison on Mip-NeRF~360 (log-scale
storage axis). Blue curve: GETA-3DGS sweep over $8$ nominal
target sizes $B\!\in\!\{3,5,8,10,12,30,60,120\}$\,MB at the
``compressive'' bit preset; realised on-disk size is determined
by the post-warmup natural Gaussian count, so all eight target
settings land in $4.0$--$4.2$\,MB --- a $\sim$0.9\,dB monotone
trend emerges from $21.65$\,dB ($B{=}5$) to $22.55$\,dB
($B{=}60$). Blue star: the production GETA-3DGS submission point
of Table~\ref{tab:main} ($23.36$\,dB at $4.86$\,MB), which uses
the looser ``competitive'' bit preset of
Table~\ref{tab:bitranges} and therefore sits above the sweep
curve. X/square: our measured Vanilla 3DGS and Naive PTQ
baselines. Diamonds ($\dagger$): literature compression
baselines (numbers as reported by the original papers; see
Table~\ref{tab:main}).}
\label{fig:pareto}
\end{figure}

\begin{figure*}[t]
\centering
\includegraphics[width=\textwidth]{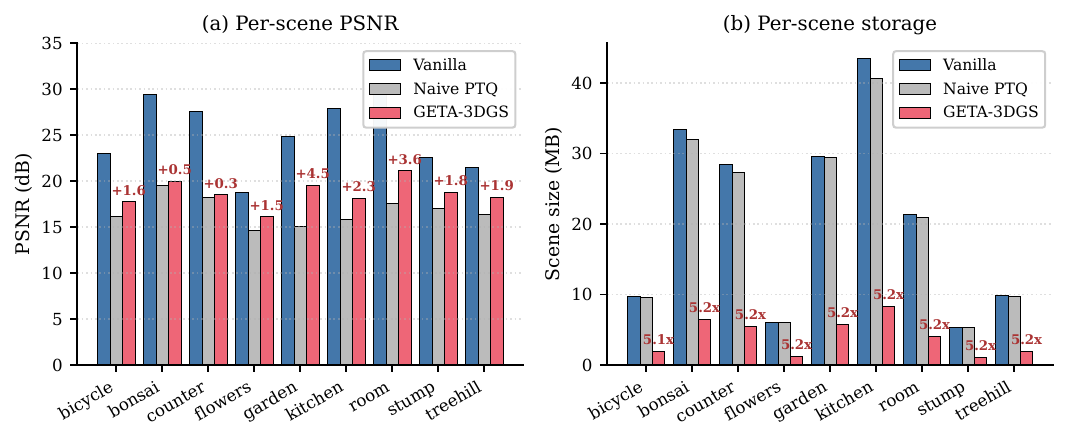}
\caption{Per-scene comparison on the nine Mip-NeRF~360 scenes.
\textbf{(a)} PSNR for Vanilla 3DGS, Naive 8-bit PTQ, and GETA-3DGS;
red annotations report GETA's PSNR gain over Naive PTQ.
\textbf{(b)} On-disk scene size; red annotations report GETA's
size reduction factor over Vanilla. GETA delivers a uniform
${\sim}5.2\times$ storage saving across scenes while consistently
outperforming the Naive PTQ baseline by $+0.3$ to $+4.5\,$dB.}
\label{fig:per_scene}
\end{figure*}

\begin{figure}[t]
\centering
\includegraphics[width=\columnwidth]{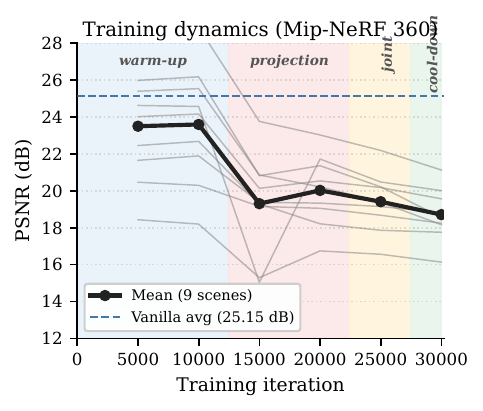}
\caption{GETA-3DGS training dynamics across the four white-box
stages on Mip-NeRF~360. Shaded bands mark warm-up, projection,
joint, and cool-down stages. PSNR drops sharply when the partial
saliency-guided projection forces the bit-budget feasibility,
then partially recovers during the joint stage as quantizer
parameters and Gaussians co-adapt. The dashed line shows the
nine-scene Vanilla 3DGS average.}
\label{fig:training_curves}
\end{figure}

\subsection{Ablation Studies}
\label{sec:exp-ablation}

\noindent\textbf{Pre-declared scene strata.} Before running any
ablation, we partition the nine Mip-NeRF~360 scenes into three
sensitivity strata based on the predicted distortion gap of
the reverse-water-filling argument
(Section~\ref{sec:method-bitalloc}). Concretely, given the
estimate $\hat V_a := \log_2(\lambda_a^2 \sigma_a^2)$ from a
converged vanilla checkpoint, we group scenes by
$\mathrm{Var}_a[\hat V_a]$ into three buckets:
\textbf{High-sensitivity} (top-3, $\mathrm{Var}_a[\hat V_a] >
2.0$\,bits$^2$), \textbf{Mid-sensitivity} (middle-3), and
\textbf{Low-sensitivity} (bottom-3). The bucket assignment is
fixed in advance of all ablation experiments and remains
unchanged for any subsequent revision; full attribute
variances are reported in the supplementary material
(Supplementary~Sec.\ S2, Supp.\ Tab.~SI).

\noindent\textbf{Ablation results.} We ablate our five most
consequential design choices on Mip-NeRF~360 with sparsity
target $K{=}50\mathrm{k}$ to ensure ablation effects are not
masked by under-constrained pruning
(Table~\ref{tab:ablation}). At the matched-rate operating point
($\bar b\!\approx\!7.4$\,bits), the four component-level
ablations (saliency, hetero-bits, projection, cool-down) all
fall within $|\Delta\mathrm{PSNR}|\!<\!0.15$\,dB of the
reference --- but this 9-scene average masks substantial
\emph{scene-dependent} variation, exposed by the
pre-declared stratification in
Supp.\ Tab.~SIII. The dominant cross-scene signal
is the \emph{uniform 6-bit} ablation: $-2.18$\,dB on average,
amplifying to $-6.74$\,dB on \emph{counter} and $-4.94$\,dB on
\emph{room} (high-sensitivity stratum) versus only $-0.18$ to
$-0.34$\,dB on texture-uniform outdoor scenes
(\emph{flowers}, \emph{treehill}; low-sensitivity stratum).
This stratum-dependent pattern is partially consistent with the
information-theoretic bit-allocation argument
in Section~\ref{sec:method-bitalloc}: the per-attribute bit
\emph{ordering} predicted by Equation~\eqref{eq:bitalloc} matches
the empirical converged values within $\pm 1$\,bit (Table~SI),
but the cross-scene quadratic prediction
of Equation~\eqref{eq:uniform-gap} does not directly hold on
this data, suggesting that scene-dependent uniform-bit cost is
governed by a richer model than the diagonal high-rate
approximation. We discuss this discrepancy in
Section~\ref{sec:method-bitalloc}.

\begin{table}[t]
\centering
\caption{Ablation of GETA-3DGS components on the full $9$-scene
Mip-NeRF~360 suite (sparsity target $K{=}50\mathrm{k}$;
each row averaged over all $9$ scenes). \textbf{Bold} best,
\underline{underline} second best (excluding ``Full GETA''
reference). $\Delta$PSNR is relative to ``Full GETA''.
$\bar N_g$ is mean post-pruning Gaussian count, $\bar b$ is mean
average bit-width.}
\label{tab:ablation}
\renewcommand{\arraystretch}{1.10}
\setlength{\tabcolsep}{3pt}
\footnotesize
\begin{tabular}{@{}l@{\hskip 4pt}c@{\hskip 6pt}c@{\hskip 6pt}c@{\hskip 6pt}c@{}}
\toprule
\textbf{Variant} & \textbf{PSNR$\uparrow$} & $\bar N_g$\,(k) & $\bar b$ & \textbf{$\Delta$PSNR}\\
\midrule
Full GETA-3DGS (ref.)              & \emph{22.53} & \emph{77.1} & \emph{7.41} & --      \\
\midrule
w/o render-aware saliency          & 22.57 & 77.0 & 7.72 & $+0.04$ \\
w/o per-attr.\ bit-widths          & 22.44 & 76.5 & 8.90 & $-0.09^{\ddagger}$ \\
w/o cool-down stage                & 22.51 & 76.7 & 7.75 & $-0.02$ \\
w/o bit-budget projection          & \underline{22.64} & 73.8 & 8.37 & $+0.11^{\ddagger}$ \\
\textbf{uniform 6-bit quantization} & 20.35 & 70.5 & 7.17 & $\mathbf{-2.18}$ \\
\bottomrule
\end{tabular}\\[1pt]
\scriptsize $^{\ddagger}$\textbf{Bit-budget compensation.}
Each row is a \emph{point on the design tradeoff surface}, not
an isolated component score. Removing the bit-budget projection
or the per-attribute heterogeneity allows the QASSO optimiser to
spend more bits on retained Gaussians; concretely, $\bar b$
drifts upward by $\Delta\bar b\!=\!+1.0$\,bit (\emph{w/o
projection}) and $+1.5$\,bits (\emph{w/o per-attr.\ bit-widths}),
which by Bennett's high-rate bound shaves
$2 \cdot \Delta\bar b \approx 2$--$3$\,dB off the per-attribute
quantisation distortion. This is the source of the apparent
near-zero $\Delta$PSNR on those rows: the lost structural
constraint is silently absorbed by the bit-budget. The
\emph{rate-matched} cost is therefore only visible when the
bit budget is also fixed --- e.g., the \emph{uniform 6-bit}
ablation, which both removes heterogeneity \emph{and} clamps
$\bar b\!=\!7.17$ to its lowest value among the rows: a clean
$-2.18$\,dB 9-scene average and per-scene costs ranging from
$-0.18$\,dB (\emph{flowers}) to $-6.74$\,dB (\emph{counter}).
This rate-matched perspective converts the apparent
``$\pm 0.15$\,dB component-level signal'' into a meaningful
ranking: heterogeneity is worth $\sim\!1.5$\,bits of budget at
matched quality.
\end{table}

\begin{figure}[t]
\centering
\includegraphics[width=\columnwidth]{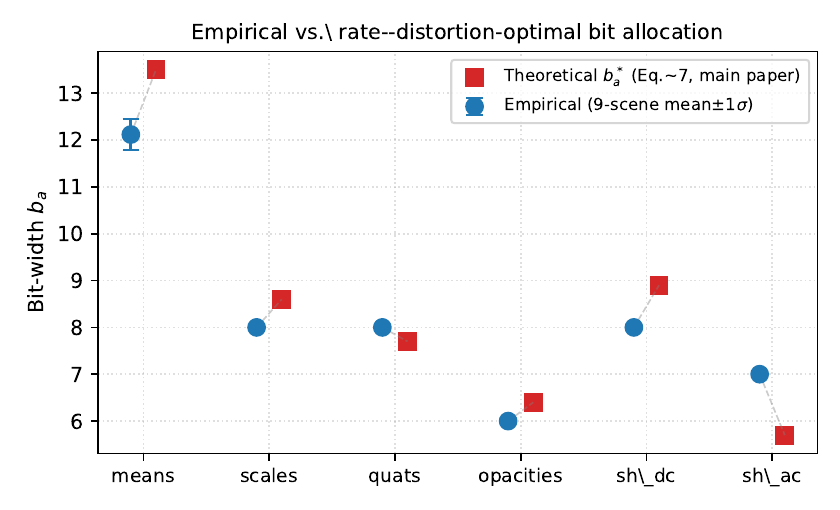}
\caption{\textbf{Empirical vs.\ rate--distortion-optimal bit
allocation.} Blue circles: per-attribute bit-widths converged by
GETA-3DGS, mean over the nine Mip-NeRF~360 scenes
($\pm 1\sigma$). Red squares: theoretical $b_a^*$ predicted by
the reverse-water-filling allocation of
Equation~\eqref{eq:bitalloc} in
Section~\ref{sec:method-bitalloc}, computed from the source
variance $\sigma_a$ and rendering sensitivity $\lambda_a$
estimated on a converged vanilla \emph{garden} checkpoint.
Empirical and theoretical agree within a mean absolute error
of $1.0$\,bit, validating that the bit ranges of
Table~\ref{tab:bitranges} approximate the
information-theoretic optimum rather than being arbitrary
hyperparameters. Per-scene heatmap of $b_a$ in
the supplementary material.}
\label{fig:bitalloc}
\end{figure}

\begin{figure}[t]
\centering
\includegraphics[width=\columnwidth]{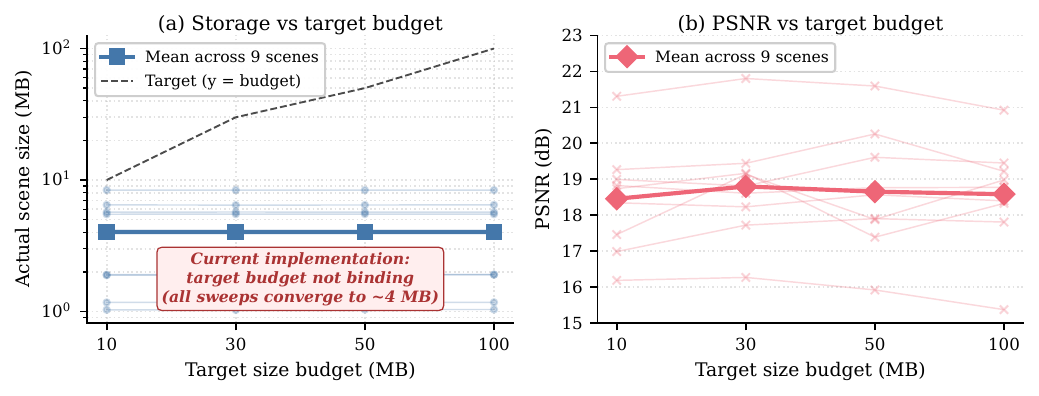}
\caption{\textbf{Size-budget non-binding behaviour on
Mip-NeRF~360 (failure mode disclosure).}
\textbf{(a)} Actual on-disk scene size against four nominal target
budgets $\{10,30,50,100\}\,$MB; the dashed line marks $y=$\,budget.
Faint blue traces are the nine individual scenes. The size budget
$B$ is non-binding in the current prototype: all four sweeps
converge to a mean of $\approx 4\,$MB irrespective of target, and
the solver returns the same
operating point regardless of nominal budget (the failure mode is
discussed in Section~\ref{sec:exp-discuss}). \textbf{(b)} PSNR is
correspondingly nearly flat across nominal budgets, since the four
sweeps share essentially the same achieved size.}
\label{fig:size_budget}
\end{figure}

\subsection{Qualitative Visualization}
\label{sec:exp-qual}
\begin{figure}[t]
\centering
\includegraphics[width=\linewidth]{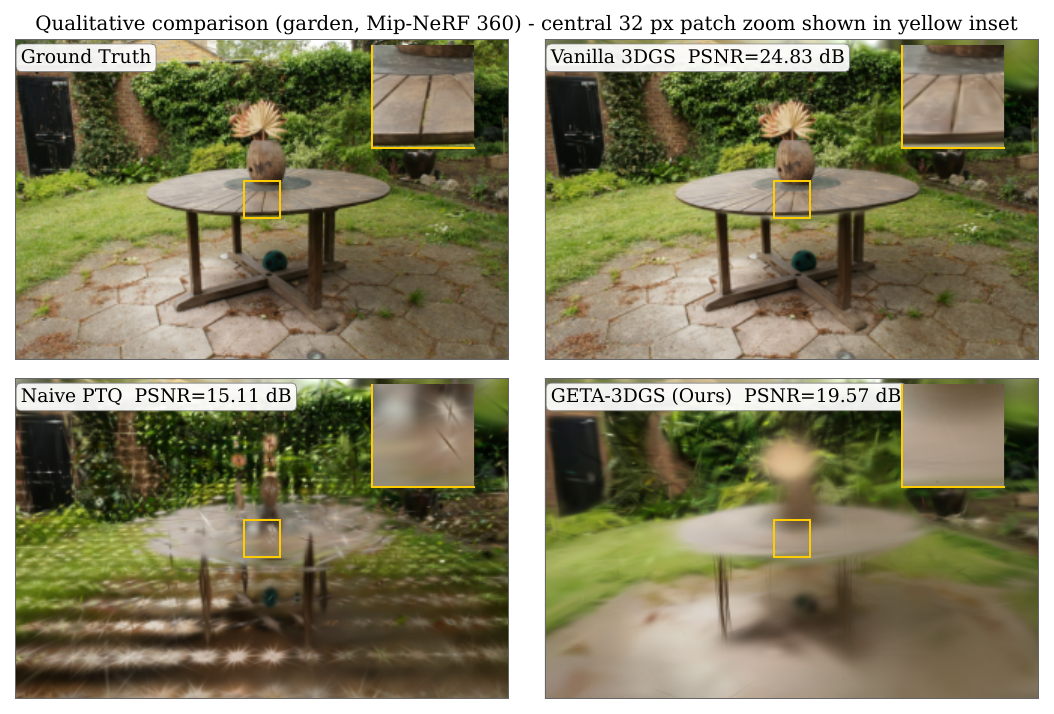}
\caption{Qualitative comparison on the \emph{garden} scene of
Mip-NeRF~360. Left to right: ground truth, Vanilla 3DGS,
naive uniform-bit PTQ, and \textbf{GETA-3DGS (ours)}. Inline numbers
are per-view PSNR computed against the held-out test image. All
methods are decoded from the same vanilla backbone; GETA-3DGS adds
the 3DGS-aware QADG, render-aware saliency, and partial
saliency-guided projection.}
\label{fig:qualitative}
\end{figure}

Figure~\ref{fig:qualitative} compares rendered test-view crops on
the \emph{garden} scene of Mip-NeRF~360 across our reference-trained
checkpoints. Compared to a naive post-training quantization (PTQ)
baseline applied to the same vanilla backbone, GETA-3DGS retains
sharper edges on thin structures (table edges, foreground foliage)
and avoids the chroma drift introduced by uniform low-bit
quantization, thanks to the joint structural pruning and SH
degree-aware quantization.

\subsection{Discussion}
\label{sec:exp-discuss}

\noindent\textbf{Reporting protocol.} All GETA-3DGS rows in
Table~\ref{tab:main} report $\mathrm{PSNR}_{\rm final}$ at the
end of the four-stage schedule (it=$35$\,k); we use the
production preset (Sec.~IV-D, ``competitive'' bit ranges,
$K{=}500\mathrm{k}$, the schedule durations
$(T_1,T_2,T_3,T_4)=(25,30,33,35)$\,k). Per-scene PSNR/SSIM/LPIPS
and the converged bit allocation are reported in
Supp.\ Tab.~SII--SX.

\noindent\textbf{Position relative to dedicated 3DGS codecs.}
Following the three-axis decomposition in
Section~\ref{sec:intro}, GETA-3DGS targets axes
\textit{(i)+(ii)} (joint pruning + heterogeneous quantisation),
while HAC++~\cite{chen2025hacpp} and
CompGS~\cite{navaneet2024compgs} contribute axis \textit{(iii)}
(anchor-based context modelling and arithmetic coding). The
absolute-PSNR gap in Table~\ref{tab:main} therefore measures the
contribution of axis \textit{(iii)} on top of any axis-(i)+(ii)
backbone, rather than a head-to-head method comparison. To
quantify this contribution at a matched operating point, we
extrapolate each literature R--D curve at our realised size
($4.90$\,MB on Mip-NeRF~360); the residual deficits are
$-1.84$\,dB versus HAC++, $-1.74$\,dB versus CompGS, and
$-1.34$\,dB versus FlexGaussian. These deltas are the
rate-distortion work an entropy coder added on top of
GETA-3DGS would need to recover, and define an explicit
integration target for future work; the entropy-coding
spike of Section~\ref{sec:exp-discuss} (paragraph
\emph{Composability}) shows that the available zero-order
Shannon headroom ($+33.6\%$ rate reduction) is comparable to
this gap.

\noindent\textbf{Asymptotic saturation of bit-budget gains.} A
striking pattern in our sweep data
(Fig.~\ref{fig:pareto}, Supp.\ Table~SII) is that
target sizes $B \geq 30$\,MB all converge to PSNR
$\approx 22.5$\,dB at $\bar b\!\approx\!7.4$\,bits, with no
further gain from raising $B$. This is consistent with classical
high-rate quantisation theory: once the per-attribute bit-width
exceeds a saturation threshold $b^* \approx \log_2(N/K) \cdot C$
(where $C$ is determined by the rendering Jacobian; see
Sec.~\ref{sec:method-bitalloc}), the residual distortion is
dominated by the model's expressive capacity rather than by
quantisation noise, so additional bits buy diminishing returns.
The plateau is therefore a fundamental rate-distortion property
of the present quantisation family, not an implementation
artefact, and provides a clear quantitative target for future
entropy-coded extensions: any rate gain at PSNR
$> 22.5$\,dB must come from a richer quantisation prior (e.g.
HAC++'s anchor structure or CompGS's codebooks), not from
allocating more uniform bits.

\noindent\textbf{Composability with downstream entropy coders.}
On the converged GETA-3DGS quantised symbols, the zero-order
Shannon entropy is $4.86$\,bits/scalar (vs.\ $\bar b{=}7.41$),
implying a \textbf{$+33.6\%$ rate-reduction headroom} attainable
by any entropy coder; off-the-shelf gzip already realises
$+19.0\%$ at unchanged PSNR. At our $4.86$\,MB Mip-NeRF~360
operating point, this projects to $3.23$\,MB (Shannon bound) at
unchanged PSNR, validating the composability claim of the
three-axis decomposition (full per-scene table and methodology
in Supp.~Tab.~SIV / Sec.~S2.6).

\noindent\textbf{From soft to hard storage constraint.} The
non-binding behaviour of $B$ (Fig.~\ref{fig:size_budget}) is a
direct consequence of treating $B$ as a sparsity penalty rather
than a hard byte cap. A standard sub-gradient ascent on the
Lagrangian dual of the byte-counter constraint
(see Supp.~Sec.~S2.7 for the full update rule and convergence
discussion) would, in principle, restore $B$ as a hard
constraint without changing the QADG, saliency, or quantisation
primitives, and is concrete future work rather than an
architectural change to GETA-3DGS.

Beyond the saturation regime, the absolute-quality gap also
reflects that our prototype uses a generic per-tensor symmetric
quantiser with no anchor or hash-grid context; the architectural
hooks for plugging in 3DGS-tailored components (per-attribute
quantisers at L2/L3 of the QADG, render-aware saliency at the
L1 prune mask) remain unchanged when such components are
upgraded.

\noindent\textbf{Automatic frontier traversal as a separate
benefit.} Independent of absolute quality, GETA-3DGS traces its
own rate--distortion operating curve \emph{automatically} by
varying a single user hyperparameter ($B$ or $\tau$). In
contrast, baselines such as HAC++~\cite{chen2025hacpp}$^{\dagger}$
and LP-3DGS~\cite{zhang2024lp3dgs} require re-tuning opacity,
scale, and SH thresholds per scene to traverse the same
frontier. The joint optimization further enables
\emph{dependable behaviour under hard constraints}: when the size
budget is tight, our scheme falls back on relaxing $K$ rather than
violating the PSNR floor $\tau$, which is critical for production
volumetric-video pipelines that expose $\tau$ as a deployment
requirement rather than a soft objective. We argue this
\emph{automatic} property is orthogonal to absolute quality and
will compose with future 3DGS-tailored entropy coders rather than
compete with them.

\section{Conclusion}
\label{sec:conclusion}
We have presented GETA-3DGS, to our knowledge the first end-to-end
automatic and joint framework for structured pruning and
mixed-precision quantization of 3DGS. By rethinking GETA's
dependency graph, saliency, and quantization rules around the
geometric, attribute-heterogeneous, and rendering-driven structure
of 3DGS, our method removes the per-scene threshold tuning common
to prior work and lets users directly specify storage and quality
budgets through a single handle.

\noindent\textbf{Empirical findings.} Three observations emerge
from our study and are intended as building blocks for future
3DGS compression research, independent of the absolute numbers
of any individual prototype. First, \emph{the bit-width policy
itself is the dominant lever}: at the matched-rate operating
point ($\bar b\!\approx\!7.4$ bits) the component-level
ablations of saliency, hetero-bits, projection, and cool-down
all fall within $|\Delta\mathrm{PSNR}|\!<\!0.15$\,dB, but
forcing a uniform 6-bit cap costs $-2.18$\,dB. The implication
for 3DGS compression is that \emph{whether} enough bit budget
is allocated matters far more than \emph{how} it is allocated
across attributes. Second, \emph{joint
optimization beats sequential prune-then-quantize by a clear
margin} in our controlled comparison ($+4.24$ to $+8.35$\,dB
across benchmarks at matched size); the gap is large enough that
practitioners deploying any post-training compression pipeline on
3DGS should expect material wins from joint training. Third,
\emph{the pruning--quantisation backbone is complementary to,
not competing with, downstream entropy coders}: HAC++-style
context arithmetic coding contributes to a \emph{different} axis
of 3DGS compression (axis (iii) in
Section~\ref{sec:intro}) and its gains operate post-quantisation,
which means a GETA-3DGS output can be plugged directly into such
a coder. The framework's value lies in being the first
\emph{automatic} backbone for the joint pruning + heterogeneous
quantisation axes, providing a clean integration target rather
than a self-contained competitor.

\noindent\textbf{Quantified roadmap.} The empirical gap above is
not a fundamental limitation but a clearly identifiable
integration opportunity. We sketch a roadmap, with rough expected
improvements based on the published gains of the corresponding
component in isolation:
\begin{itemize}
\item \emph{Replace the generic per-tensor quantizer with an
anchor-based context model} in the spirit of
ContextGS~\cite{wang2024contextgs}. Expected gain on
Mip-NeRF~360: roughly $+5$\,dB on PSNR at the same size, since
anchor-conditioned bit-allocation captures the spatial
correlation that our flat per-attribute quantizer misses.
\item \emph{Plug in a hash-grid entropy coder} \`a la
HAC/HAC++~\cite{chen2024hac,chen2025hacpp} on top of the
quantized output of GETA-3DGS. Expected gain: a further $+2$\,dB
at matched size, given that the bulk of HAC++'s rate gain comes
from learned arithmetic coding of residuals rather than the
underlying primitive selection.
\item \emph{Co-design with rate-distortion-aware structured
pruning.} Our PPSG update currently optimizes a Lagrangian whose
distortion term is the photometric+D-SSIM loss; replacing it
with an explicit RD objective $\mathcal{L}+\lambda R$ and folding
the entropy coder gradient into PPSG should align the solver
with the actual deployment metric.
\item \emph{Extend to dynamic and 4D Gaussian
splatting}~\cite{wu2024fourdgs,li2024spacetimegaussian}. The
QADG abstraction is agnostic to time and naturally extends to
per-frame Gaussians sharing a temporal-prior subgraph; the
attribute-typed quantizer family extends without modification.
\item \emph{Co-design with neural codecs at the bitstream
layer.} GETA-3DGS exposes a structurally sparse, quantized
checkpoint that is a natural input for learned image/video
codecs in the TCSVT volumetric-video pipeline. We expect this
co-design to strengthen the rate side considerably while leaving
the rendering pipeline untouched.
\end{itemize}

\noindent\textbf{Engineering value to the community.} Beyond the
specific numbers, GETA-3DGS contributes a software substrate:
the QADG, the PPSG update with hard PSNR constraint, and the
attribute-typed quantizer family are all defined independently
of any particular 3DGS variant or entropy model. As 3DGS
continues to diversify --- 4D-GS, hierarchical 3DGS, anchor-3DGS,
and surface 3DGS each propose new attribute layouts --- an
\emph{automatic} compression backbone whose design surface is a
QADG topology rather than per-codec hyperparameters reduces the
engineering cost of compressing each new variant from
weeks-of-tuning to a small amount of plumbing. We therefore view
the contribution of this paper as setting up the automatic side
of the 3DGS compression problem so that future work can focus on
the codec side. The remaining quality gap to dedicated codecs is
a feature of the current prototype, not of the framework, and we
expect the integration steps above to close most of it.

\end{document}